%% file: SGA_arXiv_2026.tex
\documentclass[review,10pt]{JMtemplate}

\usepackage{lastpage}
\usepackage[utf8]{inputenc}
\usepackage[T1]{fontenc}
\usepackage{amsmath}
\usepackage{amsfonts}
\usepackage{amssymb}
\usepackage{booktabs}
\usepackage{multirow}
\usepackage{graphicx}
\usepackage{microtype}
\usepackage[table]{xcolor}
\usepackage{subcaption}
\usepackage{makecell}
\usepackage{caption}
\usepackage{enumitem}
\usepackage{marvosym}
\usepackage{algorithm}
\usepackage{algorithmic}
\graphicspath{{media/}}

\usepackage[
    colorlinks=true,
    linkcolor=blue,
    urlcolor=blue,
    citecolor=green,
    filecolor=magenta
]{hyperref}
\usepackage{textcomp}

\begin{document}
\begin{frontmatter}
\title{SGA: Uncertainty Quantification for Multi-Step Forecasting in Time Series Foundation Models}

\author{\textbf{Xin-Yu Hu}\textsuperscript{\rm 1,2} \quad
\textbf{Shuang Liang}\textsuperscript{\rm 1,2} \quad
\textbf{Cheng Feng}\textsuperscript{\rm 3,4} \quad
\textbf{Shao-Qun Zhang}\textsuperscript{\rm 1,2,4,\Letter} \\[0.3em]
\small \textsuperscript{1} National Key Laboratory for Novel Software Technology, Nanjing University, China.\\
\small \textsuperscript{2} School of Intelligent Science and Technology, Nanjing University, China.\\
\small \textsuperscript{3} Siemens Data and AI Research, Beijing, China.\\
\small \textsuperscript{4} Nanjing University -- Siemens Joint Research Center on Industrial AI, Suzhou, China.\\
\small \texttt{ zhangsq@lamda.nju.edu.cn }
}

\begin{abstract}
The recent emergence of Time Series Foundation Models (TSFMs) has significantly advanced multi-step forecasting performance, enabling accurate predictions over extended future horizons. However, existing TSFMs often suffer from significantly inherent uncertainty, which typically manifests as derived forecast branches emerging at each time step and spreading to subsequent steps; different forecast branches often exhibit varying forecasting performance, thereby undermining the credibility of TSFM forecasts. In this paper, we propose the Slicing-Graphing-Alignment (SGA) method to quantify the uncertainty of multi-step TSFM forecasts. The proposed SGA first characterizes the topology of all potential forecast branches using a directed acyclic graph, such that the graph complexity bounds the uncertainty of multi-step forecasts, and then precisely measures the graph complexity by integrating both topological information and TSFM-inherent stochasticity. Experimental results conducted on 11 TSFMs and 27 datasets demonstrate that (i) SGA achieves the best performance when ranking predictive errors with uncertainty estimates; (ii) SGA works with a more extensive and more precise sampling coverage than those of existing UQ methods, deriving a quantification mechanism fundamentally different from those of established ones; and (iii) larger model scales of TSFMs correlate with lower uncertainty estimates of multi-step forecasts, suggesting another empirical scaling law for uncertainty quantification of multi-step TSFM forecasts.

\textit{Key words:} Time Series Foundation Model, Multi-Step Forecasting, Uncertainty Quantification, Slicing-Graphing-Alignment
\end{abstract}
\end{frontmatter}

\begin{figure*}[t]
    \centering
    \includegraphics[width=1\linewidth]{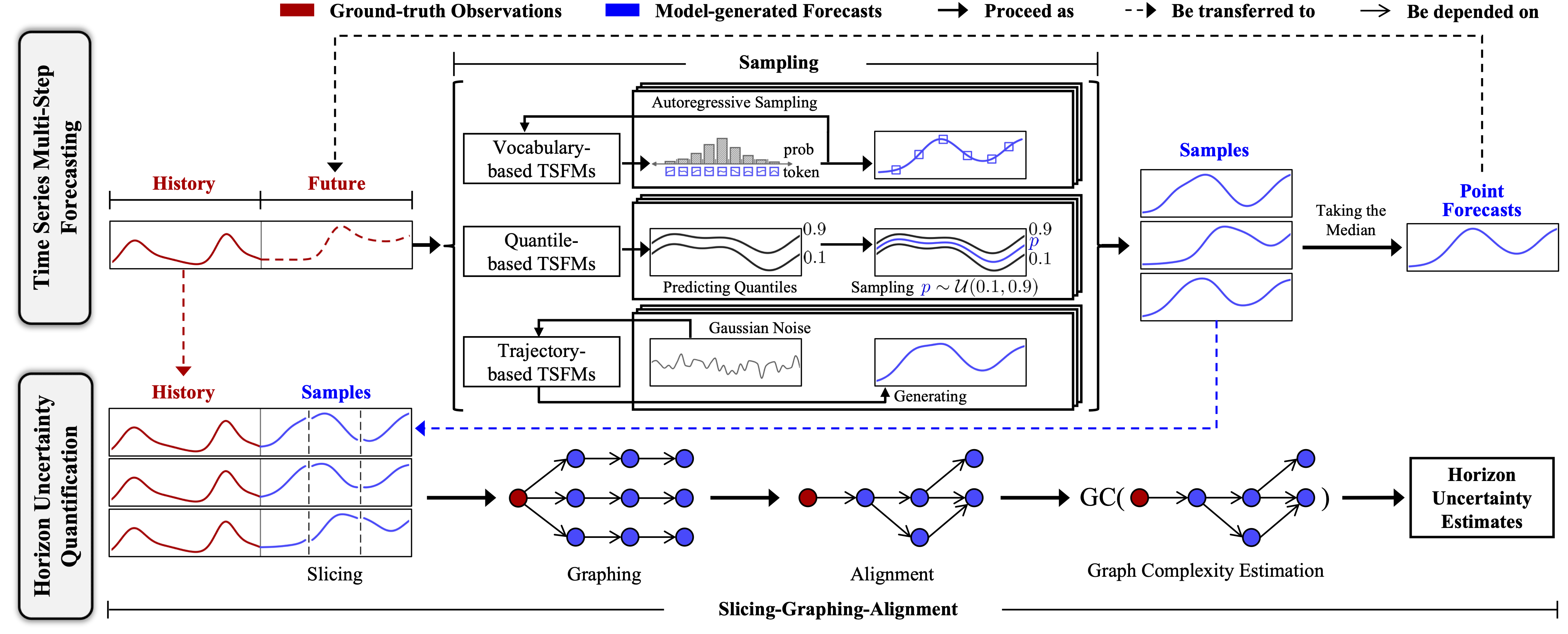}
    \caption{Workflow of time series multi-step forecasting and horizon uncertainty quantification.}
    \label{fig:overview}
\end{figure*}

\section{Introduction}  \label{sec:intro}
Many real-world applications, such as quantitative trading~\cite{sezer2020financial}, weather forecasting~\cite{shi2015weather}, and industrial power-load forecasting~\cite{hong2016electric}, can be modeled as multi-step forecasting tasks, which have been studied for decades and remain a hot topic~\cite{bentaieb2012multistep,ansari2025chronos}. Recently developed Time Series Foundation Models (TSFMs) have demonstrated superior predictive accuracy in multi-step forecasting compared with many previous approaches~\cite{ansari2024chronos,ansari2025chronos,liu2025sundial}, marking a significant milestone in the field. 

Apart from accurate multi-step forecasts, their credibility is often required by the developers for risk-benefit trade-offs, especially when they apply TSFMs in high-stakes scenarios. For example, traders frequently encounter a critical dilemma when TSFMs forecast sustained price growth for the next week without providing any credibility measures such as uncertainty; ignoring this forecast means missing out on potential profits, whereas trusting it to trigger an excessively large trading position risks substantial financial losses~\cite{fu2025financial}. Such credibility concerns primarily stem from model uncertainty arising from the training process and aleatoric uncertainty inherent in the data~\cite{chinta2026probfm}. Therefore, it is necessary and important to quantify the uncertainty of the multi-step forecasting process performed by TSFMs to assess the credibility of their forecasts. However, little attention has been paid to this topic.

There are two intuitive perspectives on extending established Uncertainty Quantification (UQ) methods to the UQ of multi-step forecasting. From the perspective of the processes concerned, one can leverage the UQ methods developed for the one-step forecasting process. These methods typically quantify the uncertainty of a single-step forecast using the width of a quantile-based interval~\cite{adler2026calibration}, which is either directly produced by the model~\cite{ansari2025chronos} or empirically estimated from multiple forecast samples~\cite{ansari2024chronos}. Hence, it is intuitive to extend such approaches by averaging the uncertainty estimates across individual time steps. From the perspective of the models concerned, one can adapt the UQ methods developed for Large Language Models (LLMs). These methods typically model generated outputs as sequences of tokens and exploit information-theoretic metrics~\cite{fomicheva2020ppl} of a single sequence or diversity-based measures of multiple sampled sequences~\cite{farquhar2024detecting} to quantify uncertainty. Thus, an intuitive approach is to model a multi-step forecast as a sequence of forecast slices, each comprising one or more consecutive time steps. Nevertheless, these two intuitive extensions struggle to precisely describe the potential branches of multi-step forecasts. Such branches emerge at each time step due to the intrinsic uncertainty of TSFMs~\cite{chinta2026probfm} and spread to the subsequent steps, forming a complex topology we term the forecasting space, which encompasses all possible multi-step forecasts. Therefore, achieving a precise characterization of potential branches during multi-step forecasting is fundamental to the UQ of this process, yet remains challenging.

Built upon this recognition, we propose the Slicing-Graphing-Alignment (SGA) method to quantify the uncertainty of the multi-step forecasting process. The slicing stage samples multiple forecasts and segments each into slices spanning one or more consecutive time steps, where slices covering the same temporal span represent potential forecast branches. The graphing stage constructs a Directed Acyclic Graph (DAG), where nodes and edges separately indicate slices and temporal dependency between consecutive slices, thereby roughly describing the topology of the forecasting space. The alignment stage aligns slices using the Dynamic Time Warping (DTW) method~\cite{sakoe1978dtw}, measures their similarity based on the resulting DTW distance, and merges similar slices to reveal the intrinsic topology of the forecasting space. The uncertainty of multi-step forecasting is then quantified by the forecasting space complexity, which can be approximated by graph complexity. We construct graph complexity by integrating both topological information and TSFM-inherent stochasticity. Experimental results across 11 TSFMs and 27 datasets demonstrate that (i) SGA significantly outperforms its contenders in ranking predictive errors with uncertainty estimates; (ii) SGA covers the forecasting space more extensively and precisely than existing methods in sampling, leading to a quantification mechanism inherently distinct from those of established ones; and (iii) models with larger scales tend to exhibit lower uncertainty estimates of multi-step forecasts, indicating another empirical scaling law for UQ of multi-step TSFM forecasts.

The rest of this paper is organized as follows. Section~\ref{sec:related_work} reviews related work. Section~\ref{sec:huq} formally introduces the SGA method. Section~\ref{sec:experiments} conducts experiments to validate the effectiveness of SGA and investigate the relation between model scales and uncertainty estimates in multi-step TSFM forecasting. Section~\ref{sec:conclusions} concludes this work.

\section{Related Work}  \label{sec:related_work}
Current TSFMs can be roughly divided into three types according to their output formats, involving vocabulary-based, quantile-based, and trajectory-based models. Vocabulary-based models like Chronos~\cite{ansari2024chronos} produce a probability distribution over a frozen vocabulary, where each token in the vocabulary represents a forecast value at one time step. A multi-step forecast is generated by autoregressively sampling from the predicted distributions, and multiple such forecasts are averaged to obtain the point forecast. Quantile-based models, including TimesFM-2.5~\cite{das2024timesfm}, Chronos-2~\cite{ansari2025chronos}, and Timer-S1~\cite{2026timers1}, output a matrix of forecasts, the row and column of which separately indicate the quantile level and future time step, and the median ($50\%$ quantile) is typically used as the multi-step point forecast. Trajectory-based models such as Sundial~\cite{liu2025sundial} and Aurora~\cite{wu2025aurora} map a sample from a standard multivariate Gaussian distribution to a single multi-step forecast, termed a trajectory, and obtain the point forecast by averaging multiple trajectories.

Existing UQ methods of LLMs can be broadly divided into three categories, involving reflexive-based, information-based, and diversity-based methods~\cite{vashurin2025benchmarking}. Reflexive-based methods~\citep{kadavath2022ptrue,xiong2024can} prompt the LLM to judge the uncertainty of its own generation, making them unsuitable for adaptation to the UQ of multi-step forecasting, since TSFMs cannot be prompted in the same manner. Information-based and diversity-based UQ methods model LLM generations as sequences, then quantify uncertainty via information-theoretic metrics~\cite{fomicheva2020ppl,malinin2021mcse} and heuristic diversity metrics~\cite{lin2024generating,farquhar2024detecting,duan2024sar}, respectively. Thus, an intuitive adaptation is to model a multi-step forecast as a sequence of forecast slices, each comprising one or more consecutive time steps. 

Existing studies primarily focus on UQ for one-step forecasting, where uncertainty is typically quantified by quantile-based interval widths~\cite{adler2026calibration} either directly produced by the model~\cite{ansari2025chronos} or empirically estimated from multiple forecast samples~\cite{ansari2024chronos}. Thus, UQ for multi-step forecasting remains underexplored.

\section{Our Methods}  \label{sec:huq}
In this section, we formulate the task of quantifying the uncertainty during multi-step TSFM forecasting processes in Subsection~\ref{subsec:huq} and then formally propose the Slicing-Graphing-Alignment (SGA) method in Subsection~\ref{subsec:sga}. We start with some useful notations. Let $[N] = \{1, 2, ..., N\}$ be an integer set for $N \in \mathbb{N}^+$, and $|\cdot|$ denotes the number of elements in a collection, e.g., $|[N]|=N$, or the width of an interval, e.g., $|[0, 1]| = 1$. 

\subsection{Horizon Uncertainty Quantification}  \label{subsec:huq}
Here, we formulate the task of quantifying the uncertainty during multi-step TSFM forecasting processes, named Horizon Uncertainty Quantification (HUQ). Provided a historical time series $\boldsymbol{x}_{1:t} = (x_1, \dots, x_t) \in \mathbb{R}^t$ of length $t\in \mathbb{N}^*$ and a forecast horizon $h \in \mathbb{N}^*$, a TSFM produces the $h$-step-ahead forecasts, denoted by $\hat{\boldsymbol{x}}_{t+1:t+h} = (\hat{x}_{t+1}, \dots, \hat{x}_{t+h}) \in \mathbb{R}^h$. Following seminal work~\cite{ansari2024chronos}, we refer to $\hat{\boldsymbol{x}}_{t+1:t+h}$ as a forecast trajectory indicated by the next $h$ timestamps, where $\hat{x}_{t+s}$ is a forecast value at time step $s$ for $s \in [h]$. The goal of HUQ is to assign an uncertainty estimate $u \in \mathbb{R}$ to the entire forecast trajectory $\hat{\boldsymbol{x}}_{t+1:t+h}$ for quantifying the credibility of multi-step TSFM forecasts.

Compared with the traditional UQ task for one-step forecasting~\cite{adler2026calibration}, the investigated HUQ task is not only more general since it takes the traditional one as a particular case once developers set $h=1$, but also more challenging because the inevitable step-wise error accumulation~\cite{bentaieb2012multistep} in multi-step forecasting often amplifies uncertainty simultaneously~\cite{girard2002gaussian}.

\subsection{Slicing-Graphing-Alignment} \label{subsec:sga}
This subsection formally proposes the Slicing-Graphing-Alignment (SGA) method for HUQ, comprising stages of slicing, graphing, and alignment. We recognize that potential forecast branches would emerge at each time step due to the inherent uncertainty of TSFMs~\cite{chinta2026probfm,dey2026distill}, with such branching spreading to subsequent steps to form a complex topology that covers all possible multi-step forecasts, which we refer to as the forecasting space. Built upon this recognition, the slicing stage first samples multiple forecasts and then segments each into several slices, each containing one or more consecutive time steps, where slices spanning the same time steps represent potential forecast branches. The graphing stage further models these slice-based forecast branches and their temporal dependency as nodes and edges in a DAG, respectively, thereby roughly describing the topology of the forecasting space. Finally, the alignment stage aligns similar nodes if their DTW distance does not exceed the pre-specified threshold $\tau \in \mathbb{R}^*$, and thus reveals the intrinsic topology of the forecasting space. The above three stages are illustrated in Figure~\ref{fig:overview} and further detailed in the following paragraphs.

\paragraph{Slicing} This stage aims to represent potential forecast branches with sliced forecasts. First, we aim to obtain $K \in \mathbb{N}^*$ forecast trajectory samples. The vocabulary-based and trajectory-based TSFMs produce a point multi-step forecast by averaging $K$ forecast trajectory samples, which can be directly retained without incurring any additional computational cost. 
The quantile-based TSFMs output a matrix $\mathbf{A} = (a_{mn})_{g\times h} \in \mathbb{R}^{g\times h}$, where $a_{mn}$ indicates the forecast value with the $m$-th smallest quantile level $q_m$ for $m \in [g]$ at time step $n \in [h]$. The largest and the smallest quantile levels are denoted as $q_{\textrm{max}}$ and $q_{\textrm{min}}$, respectively. The key idea is first to construct a horizon-level joint inverse Cumulative Distribution Function (CDF) $F:\mathbb{R} \to \mathbb{R}^h$ by linearly interpolating between the forecast values $a_{mn}$ and $a_{m+1 \ n}$ for $m \in [g - 1]$ and for $n \in [h]$. Specifically, the $s$-th component of the mapping $F$ is defined as
\[
    F_{t+s} \left(x \right) = a_{m(x)s} + \frac{\left(x-q_{m(x)} \right) \left(a_{m(x)+1\ s}-a_{m(x)s}\right) }{q_{m(x)+1}-q_{m(x)}} \ ,
\]
where $m(x)=\max\{j\in[g-1] \mid q_j\leq x\}$ identifies the interval $[q_{m(x)}, q_{m(x)+1}]$ containing $x$. Based on the constructed inverse CDF, we can perform inverse transform sampling~\citep{devroye1986sampling} by drawing a quantile level $q\in \mathbb{R}$ from the uniform distribution $\mathcal{U}(q_{\textrm{min}}, q_{\textrm{max}})$ and obtaining the forecast trajectory sample $F(q) \in \mathbb{R}^h$, with such sampling procedure repeated $K$ times to obtain $K$ samples. The sampling processes for three types of TSFMs are illustrated in Figure~\ref{fig:overview}. Next, we segment the $k$-th sampled forecast trajectory $\hat{\boldsymbol{x}}_{t+1:t+h}^k$ into a sequence of slices $\boldsymbol{b}_{1:n}^k = (b_1^k, \ldots, b_n^k)$, where each slice $b_j^k = (\hat{x}_{t+l_s (j-1)+1}^k, \dots, \hat{x}_{t+l_s j}^k)$ contains $l_s \in \mathbb{N}^*$ consecutive forecast values for $k \in [K]$, $n \in \mathbb{N}^*$, and $j \in [n]$. Slices with the same index $j$ represent potential forecast branches during the corresponding temporal span.

\paragraph{Graphing} Built upon the slice-based representation of potential branches during a certain temporal span, we can exploit a DAG to characterize the forecasting space. Specifically, we model the historical time series $\boldsymbol{x}_{1:t}$ as the root, and each slice $b_j^k$ as the subsequent node. Next, directed edges indicating the temporal dependency are added from the root to the initial slice $b_1^k$, and from the slice $b_j^k$ to the consecutive one $b_{j+1}^k$ for $k \in [K]$ and $j \in [n-1]$.

\paragraph{Alignment} This stage aligns similar nodes in the constructed DAG for revealing the intrinsic topology of the forecasting space. Two nodes are considered similar if their corresponding slices share the same index $j$, and the DTW distance of these two slices is below a threshold $\tau \in \mathbb{R}^*$. We set $\tau$ by leveraging the seasonal-scaled error~\cite{hyndman2018forecasting} as
\[
    \tau = \frac{\lambda}{t-S}\sum_{i=1}^{t-S}{|}x_i-x_{i+S}| \ ,
\]
where $\lambda \in \mathbb{R}^*$ is the threshold coefficient and $S\in [t-1]$ is seasonality parameter~\cite{ansari2024chronos} inherent in most datasets for TSFMs. Such a setting roughly matches the similarity threshold scale to that of MASE, which varies significantly across different time series or slices, thereby greatly simplifying the tuning of $\tau$. After merging similar nodes, one can obtain an aligned DAG $G=(V, E)$, where $V$ and $E$ denote the collections of vertices and edges, respectively.

\paragraph{Graph Complexity Estimation} Given the aligned DAG $G$, we construct graph complexity by integrating both TSFM-inherent stochasticity and topological information. The key idea of leveraging TSFM-inherent stochasticity is to estimate the time-step-level entropy as
\[
     U(\hat{x}_{t+s}^k) = -\mathbb{E}_{X\sim P_{t+s}^k} \log P_{t+s}^k(X) \ ,
\]
where $P^k_{t+s}(\cdot)$ denotes the time-step-level distribution at time step $s$ for the $k$-th forecast trajectory sample. Hence, it suffices to derive this time-step-level distribution. Since vocabulary-based TSFMs directly output such a distribution, we can obtain the distribution concerned without incurring any additional computations. For quantile-based TSFMs, we use the $s$-th component of the constructed inverse CDF, i.e., $F_{t+s}$, to obtain time-step-level distribution as $P^k_{t+s}(\hat{x}) = \tfrac{\mathrm{d}}{\mathrm{d}x}F_{t+s}^{-1}(\hat{x})$. For trajectory-based TSFMs, we leverage kernel density estimation~\citep{rosenblatt1956kde} to approximate the time-step-level distribution as
\[
    P^k_{t+s}(\hat{x}) = \frac{1}{K \eta_{t+s}}\sum_{i=1}^{K}\mathcal{K}
    \left(
        \frac{\hat{x} - \hat{x}_{t+s}^i}{\eta_{t+s}}
    \right) \ ,
\]
where $\mathcal{K}(x) = (2\pi)^{-1 / 2}\exp(- x^2 / 2 )$ is the Gaussian kernel and $\eta_{t+s}\in \mathbb{R}^*$ denotes the bandwidth. Specifically, we follow Scott's rule~\cite{scott1992rule} and set $\eta_{t+s} = K^{-1/5}\sigma_{t+s}$, where $\sigma_{t+s}$ denotes the standard deviation of $\{\hat{x}_{t+s}^k\}_{k=1}^{K}$.
Next, we quantify the slice-level uncertainty as
\[
    U\left( b^k_j \right) = \frac{1}{l_s} \!\sum_{s=l_s(j-1)+1}^{l_sj} \! {U\left( \hat{x}_{t+s}^{k} \right)} \ ,
\]
where a larger $U(b^k_j)$ indicates higher TSFM-inherent stochasticity over this slice's temporal span. This value is then treated as the uncertainty of the slice's corresponding node. Now, we integrate both topological information and TSFM-inherent stochasticity by exploiting the Bonacich–Lloyd alpha-centrality~\cite{bonacich2001alpha} as
\[
    B(v) = U(v) + \alpha \!\sum_{(w,v) \in E} \! B(w) \ ,
\]
where $B(v)$ is the centrality score of node $v$, $w$ denotes any predecessor of $v$, $U(v)$ indicates the node uncertainty calculated by the aforementioned process, and $\alpha$ is the attenuation factor that is typically set to $0.1$~\cite{bucur2020epidemic}. Thus, a large value of $B(v)$ reflects not only high slice-level TSFM-inherent stochasticity but also high local topological complexity, since node $v$ receives connections from numerous predecessors and accumulates their centrality. The graph complexity is derived by summing all uncertainty-aware node centrality over the DAG $G$ as 
\[
    \textrm{GC} ( G ) = \sum_{ v \in V } B ( v ) \ .
\]
Therefore, a large value of $\textrm{GC}(G)$ reflects both high TSFM-inherent stochasticity during multi-step forecasting and a complex topology of the whole forecasting space. Algorithms~\ref{alg:graph} and~\ref{alg:GC} summarize the aforementioned SGA and graph complexity estimation procedures, respectively, where $d_\text{DTW}$ denotes the computation of DTW distance.

\begin{algorithm}[!t]
\caption{The SGA Algorithm}
\label{alg:graph}
\textbf{Input:} historical time series $\boldsymbol{x}_{1:t}$, forecast horizon $h$, sampling times $K$, distance threshold $\tau$ \\
\textbf{Output:} Graph $G=(V, E)$ \\
\textbf{Procedures:}
\begin{algorithmic} [1]
\STATE Initialize $G=(V, E)$ with root node $\boldsymbol{x}_{1:t}$
\STATE Initialize $U(\boldsymbol{x}_{1:t}) \gets 0$
\FOR {$k \in [K]$}
    \STATE Sample a forecast trajectory $\hat{\boldsymbol{x}}_{t+1:t+h}^k$
    \STATE Segment it into slices $\boldsymbol{b}_{1:n}^k = (b_1^k, \ldots, b_n^k)$
    \STATE Calculate slice-level uncertainty $U(b_j^k)$ 
    \STATE Initialize $UC(b_j^k) \gets \{U(b_j^k)\}$ for all $j \in [n]$
    \STATE $V \gets V \cup \{b_j^k\}_{j \in [n]}$ \ and \ $E \gets E \cup \{(\boldsymbol{x}_{1:t}, b_1^k)\}$
    \STATE $E \gets E \cup \{(b_j^k, b_{j+1}^k)\}_{j \in [n-1]}$
\ENDFOR
\STATE Initialize processed node set $V_p \gets \emptyset$ \COMMENT{Start alignment}
\FOR {$j \in [n]$}
    \STATE $V_p \gets V_p \cup \{b_j^k\}_{k\in [K]}$
    \STATE Initialize node set $V_a \gets \{b_j^k\}_{k\in [K]}$ for alignment
    \FOR {$v\in V_a$}
        \STATE $V_a \gets V_a \setminus \{v\}$ 
        \FOR {$w\in V_a$}
            \IF {$d_\text{DTW}(v, w)\leq \tau$}
                \STATE $UC(v) \gets UC(v) \cup UC(w)$
                \STATE Get the parent $p$ of $w$ where $(p, w) \in E$
                \STATE Get the child $c$ of $w$ where $(w, c) \in E$
                \STATE $E \gets E \cup \{(p, v), (v, c)\}$ \COMMENT{Merge nodes}
                \STATE $V_p \gets V_p \setminus \{w\}$ \ and \ $V_a \gets V_a \setminus \{w\}$ 
            \ENDIF
        \ENDFOR
    \ENDFOR
\ENDFOR
\STATE $U(v) \gets \text{Average}(UC(v))$ for all $v \in V_{p}$ 
\STATE Update $G$ to retain only nodes in $V_p\cup \{\boldsymbol{x}_{1:t}\}$
\end{algorithmic}
\end{algorithm}

\begin{algorithm}[t]
\caption{Graph Complexity Estimation}
\label{alg:GC}
\textbf{Input:} Graph $G=(V,E)$, node uncertainty $\{U(v)\}_{v \in V}$, attenuation factor $\alpha$ \\
\textbf{Output:} Graph complexity $\text{GC}(G)$\\
\textbf{Procedures:}
\begin{algorithmic}[1]
\STATE $B[v] \gets U(v)$ for all $v \in V$
\STATE $L \gets \text{TopologicalSort}(G)$ \COMMENT{Get traversal order}
\FOR {$v \in L$}
    \STATE $B[v]\gets B[v]+\alpha \sum_{(w,v) \in E} B[w]$
\ENDFOR
\STATE $\text{GC}(G) \gets \sum_{v \in V} B[v]$
\end{algorithmic}
\end{algorithm}

\section{Experiments}  \label{sec:experiments}
This section conducts experiments to answer three questions of whether and to what extent (Q1) the proposed SGA outperforms existing UQ methods in ranking predictive errors of multi-step forecasts, (Q2) SGA covers a broader forecasting space than existing UQ methods in sampling, and (Q3) model complexity of TSFMs affects their uncertainty estimates.

\paragraph{Configurations} The evaluated TSFMs can be categorized into three types according to the output formats, including (i) vocabulary-based models such as Chronos-T5 (abbreviated as C-T5) family~\cite{ansari2024chronos} that are denoted by -Tiny, -Mini, -Small, -Base, and -Large over increasing parameter sizes, (ii) quantile-based models such as Chronos-2 (denoted as C-2)~\cite{ansari2025chronos}, TimesFM-2.5~\cite{das2024timesfm} and Timer-S1~\cite{2026timers1}, and (iii) trajectory-based models like Sundial~\cite{liu2025sundial} and Aurora~\cite{wu2025aurora}. All TSFMs with various UQ methods were evaluated on 27 datasets, which come from Chronos Benchmark II~\cite{ansari2024chronos}. Experiments were conducted on NVIDIA RTX 5090 32GB GPUs $\times$4.

We extended two types of UQ methods for HUQ as contenders. First, UQ methods designed for one-step forecasting, such as Native Calibration (NC)~\cite{adler2026calibration}, are extended in HUQ by averaging their step-wise uncertainties to obtain an overall uncertainty estimate of a multi-step forecast. Second, we selected 8 representative sequence-modeling-based UQ methods developed for LLMs, including information-based methods such as Perplexity (Ppl)~\citep{fomicheva2020ppl} and Predictive Entropy (PE)~\citep{malinin2021mcse}, and the diversity-based methods such as Eccentricity (Ecc), Sum of Eigenvalues (Eig), Degree matrix (Deg) of Graph Laplacian~\citep{lin2024generating}, Semantic Density (SD)~\citep{qiu2024semantic}, Shifting Attention to Relevance (SAR)~\citep{duan2024sar}, and Semantic Entropy (SE)~\citep{farquhar2024detecting}. We adapt these methods by first treating each forecast trajectory as a sequence of forecast values, and then replacing the standard similarity measure between token sequences with a DTW-based similarity between forecast trajectories $\hat{\boldsymbol{x}}$ and $\hat{\boldsymbol{x}}'$, computed as $s(\hat{\boldsymbol{x}}, \hat{\boldsymbol{x}}') = \exp(-d_{\text{DTW}}(\hat{\boldsymbol{x}}, \hat{\boldsymbol{x}}'))$. Moreover, we consider the UQ baseline that randomly samples uncertainty estimates uniformly from the interval $[0, 1]$, denoted as ``Random (Rnd)''. Details regarding contenders are provided in Appendix~\ref{app:contenders}.

For evaluations of the performance of the point forecast trajectory, we follow~\citet{ansari2025chronos} and adopt the Mean Absolute Scaled Error (MASE) metric. Following seminal studies~\citep{farquhar2024detecting,lin2024generating}, we evaluate a UQ method via the downstream task of selective prediction~\citep{geifman2017selective}, which reflects its ability to rank predictions by uncertainty such that highly erroneous predictions are assigned high uncertainty estimates. To quantify such ability, we employ Normalized Excess Area Under the Risk-Coverage Curve (NEAURC), which is a normalized variant of Excess Area Under the Risk-Coverage Curve (EAURC)~\citep{geifman2019AURC}. The lower the NEAURC, the better the ranking performance. We bootstrap datasets 1000 times and report the mean and standard deviation (std) of NEAURC. Notably, the NEAURC of Rnd is theoretically constant at $100\%$. Appendix~\ref{app:evaluations} provides details regarding evaluations.

\begin{table*}[t]
  \centering
  \setlength{\tabcolsep}{1.5pt}
  \footnotesize
  \resizebox{\linewidth}{!}{
  \begin{tabular}{lccccccccccc}
    \toprule
    \multirow{2}{*}{\textbf{UQ}} & \multicolumn{5}{c}{\textbf{Vocabulary-based}} & \multicolumn{4}{c}{\textbf{Quantile-based}} & \multicolumn{2}{c}{\textbf{Trajectory-based}} \\
    \cmidrule(lr){2-6} \cmidrule(lr){7-10} \cmidrule(lr){11-12}
        & C-T5-Tiny                                     & C-T5-Mini                                        & C-T5-Small                                    & C-T5-Base                                     & C-T5-Large                                    & C-2-Small                                     & C-2                                           & TimesFM-2.5                                   & Timer-S1                                         & Sundial                                       & Aurora                                            \\
    \midrule
    Rnd & 100.00\textsubscript{0.00\phantom{0}}         & 100.00\textsubscript{0.00\phantom{0}}            & 100.00\textsubscript{0.00\phantom{0}}         & 100.00\textsubscript{0.00\phantom{0}}         & 100.00\textsubscript{0.00\phantom{0}}         & 100.00\textsubscript{0.00\phantom{0}}         & 100.00\textsubscript{0.00\phantom{0}}         & 100.00\textsubscript{0.00\phantom{0}}         & 100.00\textsubscript{0.00\phantom{0}}            & 100.00\textsubscript{0.00\phantom{0}}         & \underline{100.00\textsubscript{0.00\phantom{0}}} \\
    NC  & 93.72\textsubscript{11.01}                    & 100.31\textsubscript{9.74\phantom{0}}            & 94.93\textsubscript{10.75}                    & 95.22\textsubscript{11.09}                    & 97.99\textsubscript{11.81}                    & 88.23\textsubscript{10.79}                    & 89.81\textsubscript{11.59}                    & 93.94\textsubscript{12.00}                    & 96.99\textsubscript{10.70}                       & 93.93\textsubscript{11.37}                    & 100.31\textsubscript{10.35}                       \\
    Ppl & 89.33\textsubscript{11.52}                    & 87.81\textsubscript{10.24}                       & 92.21\textsubscript{12.82}                    & \underline{88.88\textsubscript{12.16}}        & 93.43\textsubscript{11.42}                    & \underline{86.91\textsubscript{10.55}}        & \underline{84.74\textsubscript{10.41}}        & \underline{87.03\textsubscript{11.81}}        & \underline{93.11\textsubscript{9.96\phantom{0}}} & \underline{93.46\textsubscript{11.14}}        & 100.14\textsubscript{10.89}                       \\
    PE  & \underline{84.20\textsubscript{11.62}}        & \underline{84.61\textsubscript{9.14\phantom{0}}} & \underline{90.89\textsubscript{12.66}}        & 89.47\textsubscript{10.44}                    & 91.77\textsubscript{11.86}                    & 91.21\textsubscript{10.53}                    & 88.80\textsubscript{11.21}                    & 88.10\textsubscript{11.64}                    & 93.91\textsubscript{10.14}                       & 96.07\textsubscript{11.91}                    & 101.83\textsubscript{10.77}                       \\
    Eig & 96.73\textsubscript{11.65}                    & 99.70\textsubscript{10.54}                       & 95.01\textsubscript{10.73}                    & 94.55\textsubscript{10.76}                    & 97.62\textsubscript{11.35}                    & 90.93\textsubscript{10.77}                    & 91.37\textsubscript{11.27}                    & 99.21\textsubscript{11.88}                    & 96.23\textsubscript{10.33}                       & 98.10\textsubscript{11.94}                    & 102.20\textsubscript{11.10}                       \\
    Ecc & 95.50\textsubscript{10.84}                    & 94.41\textsubscript{10.46}                       & 94.46\textsubscript{10.69}                    & 96.29\textsubscript{10.08}                    & 98.86\textsubscript{11.12}                    & 89.41\textsubscript{11.08}                    & 93.97\textsubscript{10.46}                    & 96.65\textsubscript{11.59}                    & 97.66\textsubscript{10.30}                       & 97.39\textsubscript{12.41}                    & 101.30\textsubscript{10.93}                       \\
    Deg & 96.28\textsubscript{11.73}                    & 99.49\textsubscript{10.41}                       & 94.30\textsubscript{10.72}                    & 94.12\textsubscript{10.80}                    & 97.45\textsubscript{11.48}                    & 91.13\textsubscript{10.70}                    & 91.52\textsubscript{11.34}                    & 99.50\textsubscript{12.08}                    & 96.28\textsubscript{10.43}                       & 97.85\textsubscript{11.91}                    & 101.99\textsubscript{11.02}                       \\
    SD  & 89.93\textsubscript{11.85}                    & 88.52\textsubscript{11.78}                       & 92.66\textsubscript{13.26}                    & 89.49\textsubscript{12.64}                    & 94.76\textsubscript{12.66}                    & 89.65\textsubscript{9.73\phantom{0}}          & 90.36\textsubscript{10.31}                    & 94.79\textsubscript{11.83}                    & 98.57\textsubscript{9.95\phantom{0}}             & 99.56\textsubscript{11.55}                    & 105.51\textsubscript{10.56}                       \\
    SAR & 86.28\textsubscript{11.60}                    & 92.69\textsubscript{10.35}                       & 91.17\textsubscript{11.52}                    & 90.73\textsubscript{10.88}                    & \underline{90.25\textsubscript{11.83}}        & 90.16\textsubscript{10.33}                    & 88.78\textsubscript{10.93}                    & 88.22\textsubscript{11.34}                    & 93.30\textsubscript{9.94\phantom{0}}             & 94.51\textsubscript{11.34}                    & 101.84\textsubscript{11.30}                       \\
    SE  & 84.81\textsubscript{11.27}                    & 84.66\textsubscript{9.37\phantom{0}}             & 91.92\textsubscript{12.97}                    & 90.94\textsubscript{10.20}                    & 92.61\textsubscript{11.30}                    & 90.46\textsubscript{10.22}                    & 87.14\textsubscript{10.77}                    & 87.34\textsubscript{11.47}                    & 93.47\textsubscript{10.19}                       & 95.05\textsubscript{11.47}                    & 102.13\textsubscript{11.10}                       \\
    SGA & \textbf{58.89\textsubscript{8.52\phantom{0}}} & \textbf{60.30\textsubscript{7.93\phantom{0}}}    & \textbf{61.23\textsubscript{7.23\phantom{0}}} & \textbf{61.33\textsubscript{8.12\phantom{0}}} & \textbf{62.20\textsubscript{9.53\phantom{0}}} & \textbf{49.81\textsubscript{8.22\phantom{0}}} & \textbf{44.87\textsubscript{6.31\phantom{0}}} & \textbf{49.75\textsubscript{7.49\phantom{0}}} & \textbf{47.76\textsubscript{6.20\phantom{0}}}    & \textbf{51.01\textsubscript{7.12\phantom{0}}} & \textbf{57.74\textsubscript{7.01\phantom{0}}}     \\
    \bottomrule
  \end{tabular}
  }
  \caption{Comparisons of the overall NEAURC of SGA and its contenders, averaged across 27 datasets and 11 TSFMs of 3 types, where bold and underlined values denote the best and second-best results, respectively.}
  \label{tab:naurc_comparison}
\end{table*}

\paragraph{Verifications on SGA} Table~\ref{tab:naurc_comparison} presents the comparisons of the overall NEAURC of SGA and its contenders, averaged across 27 datasets and 11 TSFMs of 3 types, where bold and underlined values denote the best and second-best results, respectively. There are two key observations. First, all contenders yield comparable overall performance to that of the Rnd baseline. Specifically, according to the $3\sigma$ rule in statistics~\cite{moore2017introduction}, a UQ method is considered to significantly outperform the Rnd baseline if its performance exceeds the baseline by three standard deviations, a criterion that all contenders fail to meet. Therefore, we can conclude that all contenders typically exhibit near-random ranking ability. Second, it is obvious that SGA outperforms the strongest contenders by $34.49\%$ in NEAURC and meets the $3\sigma$ rule. This observation indicates that our proposed SGA can effectively assign high uncertainty to a forecast trajectory with high MASE, achieving a favorable ranking ability, and thus answers Q1.

\begin{figure*}[t]
    \centering
    \includegraphics[width=\linewidth]{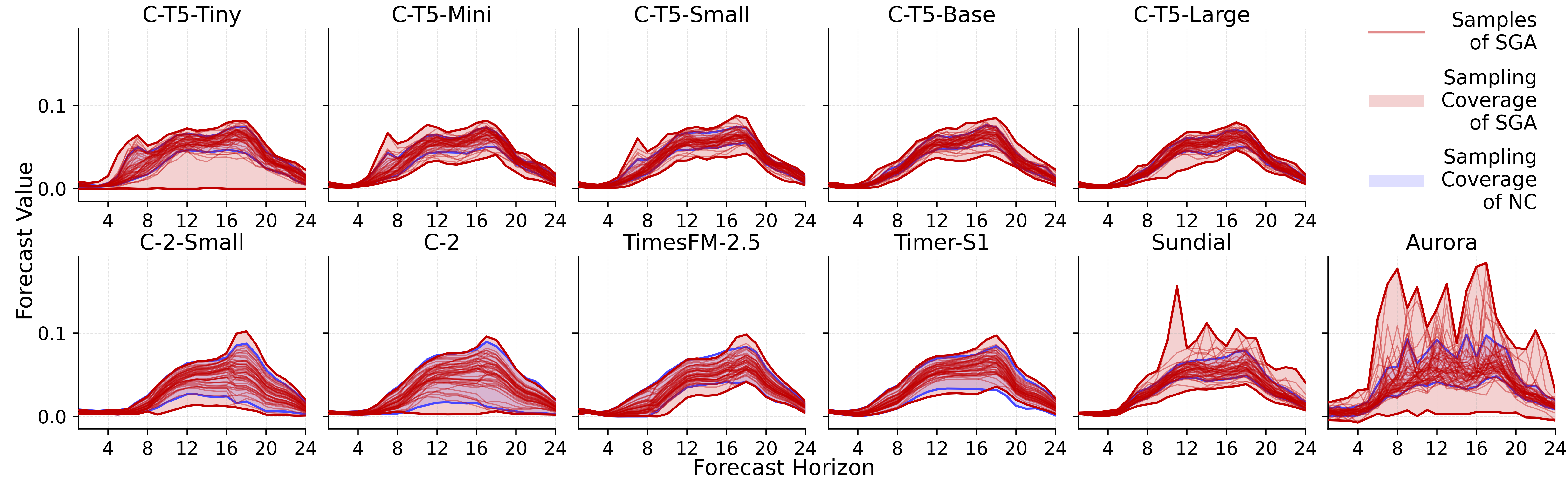}
    \caption{Visualized comparisons of the sampling coverage between SGA and NC across 11 TSFMs of 3 types.}
    \label{fig:overlap}
\end{figure*}

\begin{table}[t]
\footnotesize
\centering
\setlength{\tabcolsep}{4pt}
\begin{tabular}{cccc}
\toprule
Type & TSFM & $\tfrac{A_{\text{SGA}} \cap A_{\text{NC}} } {A_{\text{NC}}}$ & $\tfrac{ A_{\text{SGA}} \cap A_{\text{NC}}} {A_{\text{SGA}}}$ \\
\midrule
\multirow{5}{*}{Vocabulary-based} & C-T5-Tiny & 100.00 & 50.24 \\
 & C-T5-Mini & 100.00 & 49.47 \\
 & C-T5-Small & 100.00 & 48.68 \\
 & C-T5-Base & 100.00 & 48.75 \\
 & C-T5-Large & 100.00 & 48.77 \\
\midrule
\multirow{4}{*}{Quantile-based} & C-2-Small & 91.40 & 93.51 \\
 & C-2 & 91.02 & 93.86 \\
 & TimesFM-2.5 & 92.28 & 93.31 \\
 & Timer-S1 & 91.18 & 93.26 \\
\midrule
\multirow{2}{*}{Trajectory-based} & Sundial & 100.00 & 46.81 \\
 & Aurora & 100.00 & 34.64 \\
\bottomrule
\end{tabular}
\caption{Comparisons of the overall overlap area ratios for 11 TSFMs of 3 types, averaged across 27 datasets.}
\label{tab:interval_overlap_transposed}
\end{table}

\paragraph{Comparisons of Sampling Coverage} The sampling coverage $C$ of a UQ method $a$ is a set of step-wise coverage intervals $I(a)$ defined as
\[
    C(a)=\left\{ I_{t+s}\left( a \right) \right\} _{s\in [h]} \ ,
\]
where
\[
    I_{t+s}\left( a \right) \triangleq  \left[ \min_{k\in [K]} \left\{ \hat{x}_{t+s}^{k}(a) \right\} ,\max_{k\in [K]} \left\{ \hat{x}_{t+s}^{k}(a) \right\} \right] \ ,
\]
and $\hat{x}_{t+s}^{k}(a)$ denotes the $k$-th forecast sample value at time step $s$, drawn with UQ method $a$. A broader sampling coverage indicates that a method explores a more extensive region of the forecasting space. Therefore, such a method can contribute to describing the forecasting space more precisely. Figure~\ref{fig:overlap} visualizes the comparisons of the sampling coverage between SGA and its contender NC across 11 TSFMs of 3 types, with the former shown in red and the latter in blue. It is obvious that the red regions typically cover their blue counterparts, which suggests that SGA achieves a broader sampling coverage than NC. To further quantify such comparisons, we define the area $A\in \mathbb{R}$ of the sampling coverage as the sum of widths of the coverage intervals over horizon $h$, that is,
\[
    A(C(a)) = \sum_{s=1}^h{\left( \max _{k \in [K]}\left\{ \hat{x}_{t+s}^{k}(a) \right\} -\min_{k \in [K]}\left\{ \hat{x}_{t+s}^{k}(a) \right\} \right)} \ ,
\]
where a larger area indicates a broader sampling coverage. Let $A_\text{SGA}$ and $A_\text{NC}$ denote the sampling coverage areas of the sampling mechanisms inherent in SGA and its contender NC, respectively. We define the overlap in sampling coverage between SGA and NC as the sum of the intersection widths of their coverage intervals across horizon $h$, that is,
\[
    A_{\text{SGA}} \cap A_{\textrm{NC}} \triangleq \sum_{s=1}^h{|I_{t+s}\left( \text{SGA} \right) \cap I_{t+s}\left( \textrm{NC} \right) |} \ .
\]
Thus, a higher ratio $(A_{\text{SGA}} \cap A_{\text{NC}} ) / A_{\text{NC}}$ indicates that SGA envelops a larger portion of NC, while a lower ratio $( A_{\text{SGA}} \cap A_{\text{NC}}) / A_{\text{SGA}}$ implies that SGA can explore a more extensive area beyond the enclosed NC. Table~\ref{tab:interval_overlap_transposed} shows the comparisons of the overall overlap area ratios for 11 TSFMs of 3 types, averaged across 27 datasets. There are two key observations. First, all ratios in the first column exceed 90\%, indicating that SGA encompasses the vast majority of NC sampling coverage. Second, the ratios in the second column are typically low, especially for vocabulary-based and trajectory-based TSFMs, suggesting that SGA covers a much broader region beyond NC's existing sampling coverage. Overall, the aforementioned observations answer Q2.

\begin{figure}[t]
    \centering
    \includegraphics[width=0.6\linewidth]{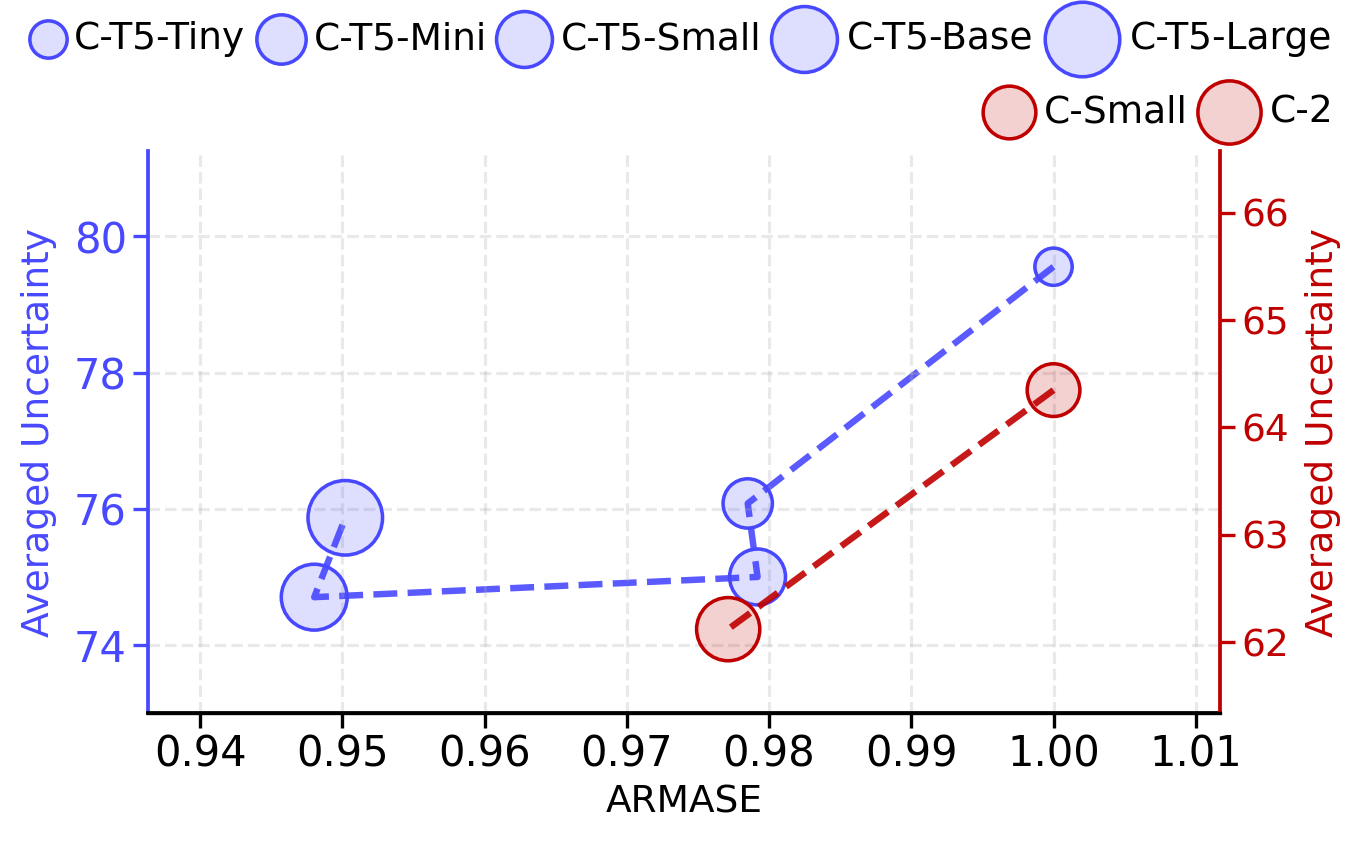}
    \caption{Plots of averaged uncertainty versus ARMASE of TSFMs over diverse scales, averaged across 27 datasets.}
    \label{fig:scale}
\end{figure}

\paragraph{Relation between Uncertainty and Scale} Figure~\ref{fig:scale} shows the plots of averaged uncertainty versus Aggregated Relative Mean Absolute Scaled Error (ARMASE)~\cite{ansari2024chronos} of TSFMs over diverse scales, averaged across 27 datasets, where computations of these metrics and results for each TSFM-dataset pair are detailed in Appendices~\ref{app:evaluations} and~\ref{app:relation}, respectively. It is observed that larger circles tend to appear in the bottom-left region, whereas smaller circles are generally located in the top-right, indicating that TSFMs with larger scales usually achieve lower predictive errors and lower uncertainty estimates. This observation reveals two conclusions. First, TSFMs with larger scales usually achieve higher forecasting accuracy, which is consistent with the scaling law of model performance~\citep{kaplan2020scaling}. Second, larger-scale TSFMs tend to produce multi-step forecasts with lower uncertainty estimates. This insight reveals an empirical scaling law of TSFM uncertainty, thus answering Q3.

\begin{figure*}[t!]
    \centering
    \includegraphics[width=\linewidth]{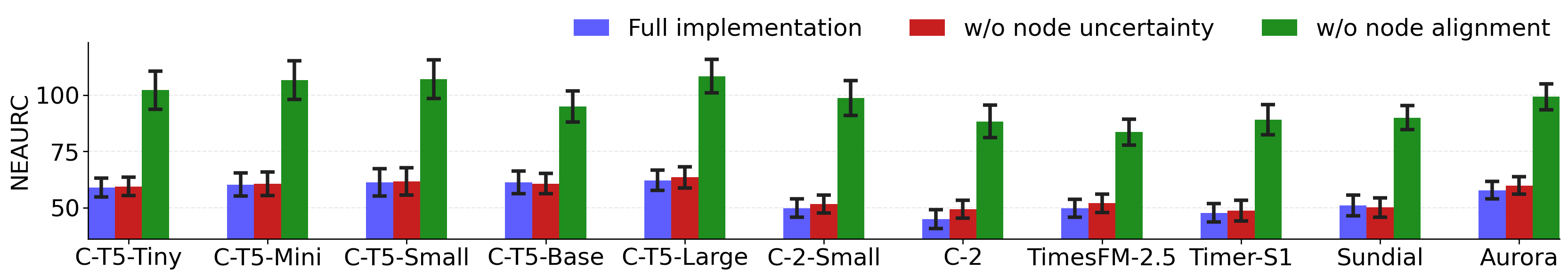}
    \caption{Ablation comparison of the overall performance of SGA on 11 TSFMs of 3 types, averaged across 27 datasets.}
    \label{fig:ablation}
\end{figure*}

\begin{figure*}[ht]
    \centering
    \includegraphics[width=\linewidth]{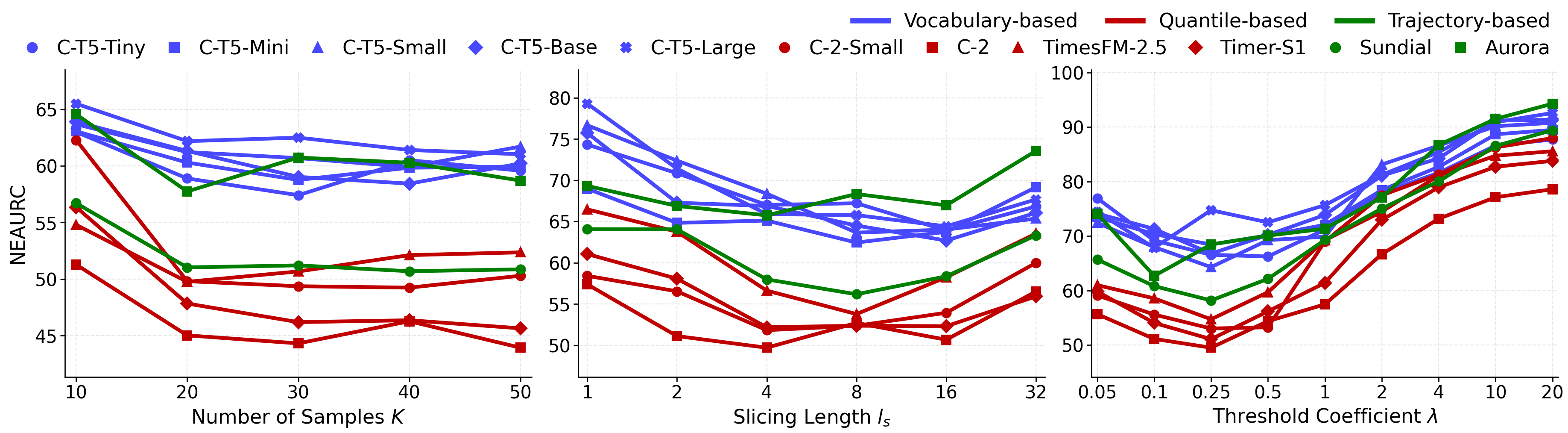}
    \caption{Impact of the number of samples $K$ (left), the slicing length $l_s$ (middle), and the threshold coefficient $\lambda$ (right) on the overall performance of SGA for 11 TSFMs of 3 types, averaged across 27 datasets.}
    \label{fig:sensitivity}
\end{figure*}

\paragraph{Ablation Analyses} Figure~\ref{fig:ablation} shows the ablation comparisons of the overall performance of SGA on 11 TSFMs of 3 types, averaged across 27 datasets. There are two key observations. First, the green bars are significantly taller than the blue bars, demonstrating that node alignment is critical to the performance of SGA. Second, the red bars are slightly taller than their blue counterparts, which suggests that when applying SGA to selective prediction, topological information contributes more significantly to the construction of graph complexity than TSFM-inherent stochasticity. Results for each TSFM-dataset pair are provided in Appendix~\ref{app:ablation}.

\paragraph{Sensitivity Analyses} Our proposed SGA involves three key hyperparameters, that is, the number of samples $K$, the slicing length $l_s$, and the threshold coefficient $\lambda$. The left, middle, and right panels of Figure~\ref{fig:sensitivity} separately show the impact of $K$, $l_s$, and $\lambda$ on the overall performance of SGA for 11 TSFMs of 3 types, averaged across 27 datasets. Since a larger $K$ leads to higher computational costs, we recommend $K=20$ for the C-2 family, TimesFM-2.5, Sundial, and Aurora, and $K=30$ for the C-T5 family and Timer-S1 to achieve a favorable trade-off between efficiency and performance. Moreover, we recommend $l_s = 4$ and $\lambda = 0.25$ for all 11 TSFMs according to the overall performance. Appendix~\ref{app:sensitivity} provides individual results on each dataset.

\section{Conclusions}  \label{sec:conclusions}
In this paper, we investigated the task of quantifying the uncertainty of multi-step forecasts in TSFMs. This work resorts to characterizing the topology of all potential forecast branches of TSFMs using a DAG, such that the graph complexity bounds the uncertainty of multi-step forecasts, and then precisely measures the graph complexity by integrating both topological information and TSFM-inherent stochasticity. Empirical results across 11 TSFMs and 27 datasets not only demonstrated the effectiveness of our method, but also revealed an empirical scaling law for multi-step forecasting uncertainty, where a larger parameter scale of TSFMs typically leads to lower uncertainty estimates.

\section*{Acknowledgments}
This research was supported by the National Science Foundation of China (62406138, 62632005) and Nanjing University -- Siemens Joint Research Center on Industrial AI.

\appendix
\onecolumn

\section*{Appendix}
This appendix provides the supplementary materials for our work ``SGA: Uncertainty Quantification for Multi-Step Forecasting in Time Series Foundation Models'', constructed according to the corresponding sections therein.

\section{Additional Implementation Details of SGA} \label{app:implement}
This section provides additional implementation details of the proposed SGA method, introducing the configurations of hyperparameters in this work.

\paragraph{Configuration of Forecasting and Sampling Parameters}
For each time series, we use the most recent history as the model input. The length of the input is set to $t=\min(l_h, 512)$, where $l_h$ denotes the length of the historical time series available before the forecast horizon.

\begin{itemize}
    \item \textbf{For Evaluating Predictive Error.}
    The point forecast used to compute MASE is obtained according to the native output form of each model. For vocabulary-based and trajectory-based models, we use 20 forecast samples to obtain the point forecast. For quantile-based models, we use the median forecast ($50\%$ quantile).

    \item \textbf{For Evaluating Uncertainty.}
    Following existing TSFM studies~\citep{ansari2024chronos,liu2025sundial}, which generate 20 forecast trajectory samples for probabilistic forecasting, we draw $K=20$ samples from each model for quantifying horizon uncertainty. In the experiments of comparisons of sampling coverage, we sample $K=50$ forecasts to showcase the difference in sampling coverage both for SGA and for NC when NC is applied to vocabulary-based and trajectory-based models.
    Sampling parameters vary for three types of TSFMs.
    For the vocabulary-output models considered in our experiments, namely the Chronos-T5 family, autoregressive sampling is performed with a temperature of $T=1.0$ and top-$k=50$.
    For quantile-output models, namely the Chronos-2 family, TimesFM-2.5, and Timer-S1, we set $q_{\textrm{min}}=0.1$ and $q_{\textrm{max}}=0.9$ to draw a quantile level $q\in \mathbb{R}$ from the uniform distribution $\mathcal{U}(0.1, 0.9)$, since $[0.1,0.9]$ is the widest quantile-level range shared by all four quantile-based models. To balance efficiency and performance, we construct an 8-step joint inverse CDF $F:\mathbb{R} \to \mathbb{R}^8$ and sample the next 8 consecutive forecast values $F(q) \in \mathbb{R}^8$. This procedure is repeated autoregressively until the complete forecast trajectory is generated.
    For trajectory-output models, namely Sundial and Aurora, we retain their default sampling configurations and directly generate $K=20$ forecast trajectory samples.
\end{itemize}

\paragraph{Configuration of SGA} 
The hyperparameters in SGA include the slicing length $l_s$ and the DTW threshold coefficient $\lambda$. For our implementation, we set $l_s=4$ and $\lambda=0.25$. 
The middle and right panels of Figure~\ref{fig:sensitivity} in Section~\ref{sec:experiments} show the impact of $l_s$ and $\lambda$ on the overall performance of SGA for 11 TSFMs of 3 types, respectively. It can be observed that SGA is relatively insensitive to small variations in $l_s$ and $\lambda$ around our default configuration and maintains stable performance across a broad range of reasonable settings, which further demonstrates the robustness of SGA.
When $l_s$ does not divide the forecast horizon $h$, we set the number of slices as $n=\lceil h/l_s\rceil$, where the symbol $\lceil z \rceil$ indicates the smallest integer not less than $z\in \mathbb{R}$. The last slice is then defined as $b_n^k=(\hat{x}_{t+l_s(n-1)+1}^k,\ldots,\hat{x}_{t+h}^k)$, which contains the remaining forecast values.


\section{Additional Experimental Details on SGA}  \label{app:experiment}
This section provides additional experimental details of the proposed SGA method.

\subsection{Evaluations} \label{app:evaluations}
\paragraph{NEAURC}
Following seminal studies~\cite{farquhar2024detecting}, we evaluate a UQ method via the downstream task of selective prediction~\citep{geifman2019selective}, where a time series prediction with higher uncertainty is expected to incur a higher predictive error. We adopt the Normalized Excess Area Under the Risk-Coverage Curve (NEAURC), a normalized variant of Excess Area Under the Risk-Coverage Curve (EAURC)~\cite{geifman2019AURC}, as a metric to evaluate the performance of a UQ method on such a downstream task. The lower the NEAURC, the better the performance. Specifically, NEAURC normalizes the traditional EAURC by calculating the ratio of the difference between the evaluated estimator and the oracle estimator to the difference between the random baseline and the oracle estimator, where the random baseline samples uncertainty estimates from $[0, 1]$ uniformly, and the oracle estimator always assigns higher uncertainty to predictions with higher predictive error. Therefore, NEAURC measures the extent to which the evaluated UQ estimator matches the ideal oracle ranking. 

Specifically, let $e_k$ denote the error of the $k$-th forecast within $n$ multi-step forecasts, which is measured by MASE in our experiments. Let $\pi$ be the ordering induced by increasing uncertainty. At the coverage level $\frac{m}{n}$ for $n\in \mathbb{N}^+$ and $m\in [n]$, the selective risk is $\text{Risk}(\frac{m}{n})=\frac{1}{m}\sum_{j=1}^{m} e_{\pi_j}$.
The AURC is defined as the average selective risk over all coverage levels
\[
    \text{AURC}
    =
    \frac{1}{n}\sum_{m=1}^{n}
    \text{Risk}\left(\frac{m}{n}\right)
    =
    \frac{1}{n}\sum_{m=1}^{n}
    \frac{1}{m}\sum_{j=1}^{m} e_{\pi_j} \ .
\]
Therefore, the calculation of NEAURC follows
\[
    \text{NEAURC} = \frac{\text{AURC} - \text{AURC}_{\text{Oracle}}}{\text{AURC}_{\text{Random}} - \text{AURC}_{\text{Oracle}}} \ ,
\]
where $\text{AURC}_{\text{Random}}$ and $\text{AURC}_{\text{Oracle}}$ denote the AURC of the random baseline and the oracle estimator, respectively.

\paragraph{ARMASE}
Following~\citet{ansari2024chronos}, we adopt the aggregated relative MASE for evaluating the overall performance of point forecasts across multiple datasets.
We first compute the relative MASE of each model on each dataset by dividing its MASE by that of a baseline model. Chronos-T5-Tiny and Chronos-2-Small serve as the baseline models for the Chronos-T5 and Chronos-2 families, respectively. Next, we aggregate relative MASE across all datasets using geometric mean, deriving the aggregated relative MASE. 
Specifically, given the historical time series $\boldsymbol{x}_{1:t}$, the ground-truth future time series $\boldsymbol{x}_{t+1:t+h}$, and the point forecast
$\hat{\boldsymbol{x}}_{t+1:t+h}$, MASE is defined as
\[
    \text{MASE} =\frac{
    \frac{1}{h}\sum_{s=1}^{h} \left|\hat{x}_{t+s}-x_{t+s}\right|
}{
    \frac{1}{t-S}\sum_{i=1}^{t-S} \left|x_i-x_{i+S}\right|
} \ ,
\]
where $S\in[t-1]$ denotes the seasonality parameter. Therefore, the calculation of ARMASE on $n$ datasets follows
\[
    \text{ARMASE} = \left(\prod_{i=1}^{n}\frac{\text{MASE}^{(i)}}{\text{MASE}_{\text{Baseline}}^{(i)}}\right)^{\frac{1}{n}} \ ,
\]
where $\text{MASE}^{(i)}$ denotes the MASE of the evaluated model on the $i$-th dataset and $\text{MASE}_{\text{Baseline}}^{(i)}$ denotes the MASE of the baseline model on the $i$-th dataset for $i\in [n]$.

\subsection{Details on Datasets} \label{app:datasets}
In this work, we conducted experiments on Chronos Benchmark II~\cite{ansari2024chronos}, containing 27 datasets spanning several application domains including energy, finance and economics, healthcare, nature, retail, and mobility and transport. These datasets cover a wide range of sampling frequencies, including 15-minute, 30-minute, hourly, daily, weekly, monthly, quarterly, and yearly frequencies. Such diversity provides a comprehensive testbed for evaluating our proposed SGA in HUQ tasks across heterogeneous forecasting regimes. For each dataset, we follow the Chronos Benchmark II evaluation protocol, where the final $h$ observations are held out as the forecast horizon and the preceding observations are used as historical time series.

\subsection{Details on UQ Contenders} \label{app:contenders}
This subsection introduces the details of the investigated contenders. 

\paragraph{UQ Methods of One-Step Forecasting}
To adapt the native calibration method~\cite{adler2026calibration}, we calculate the average of widths of central $80\%$ intervals with the $10\%$ and $90\%$ quantile forecasts from TSFMs.
For quantile-based models, the quantiles used in NC are obtained directly from model outputs. For vocabulary-based and trajectory-based models, we follow \citet{ansari2024chronos,liu2025sundial} and the official implementation of Chronos\footnote{https://github.com/amazon-science/chronos-forecasting/blob/main/src/chronos/chronos.py\#L513-L533}, and obtain each required quantile at each time step by linearly interpolating between adjacent sampled values after sorting these values. 

\paragraph{UQ Methods of LLMs}
Table~\ref{tab:llm_uq} lists 8 representative sequence-modeling-based UQ methods for LLMs that we used as contenders in our experiments, where multiple sampling means the method needs to sample multiple sequences for UQ.
All multi-sampling contenders use the same sampling procedure as SGA and obtain $K=20$ sampled forecast trajectories, ensuring a consistent sampling protocol. Our implementations of these 8 methods are based on LM-Polygraph\footnote{https://github.com/IINemo/lm-polygraph}~\citep{vashurin2025benchmarking}.

\begin{table}[htbp]
  \centering
  \footnotesize
  \setlength{\tabcolsep}{2pt}
  \begin{tabular}{cclccc}
      \toprule
      \multirow{2}{*}{Category}
      & \multirow{2}{*}{Abbr.}
      & \multirow{2}{*}{Full Name}
      & Multiple
      & Exploiting Inherent
      & \multirow{2}{*}{Literature} \\
      & & & Sampling? & Stochasticity? & \\
      \midrule
      \multirow{2}{*}{Information-based}
      & Ppl & Perplexity
      & $\times$ & $\checkmark$ & \citet{fomicheva2020ppl} \\
      & PE & Predictive Entropy
      & $\checkmark$ & $\checkmark$ & \citet{malinin2021mcse} \\
      \midrule
      \multirow{6}{*}{Diversity-based}
      & SE & Semantic Entropy
      & $\checkmark$ & $\checkmark$ & \citet{farquhar2024detecting} \\
      & SAR & Shifting Attention to Relevance
      & $\checkmark$ & $\checkmark$ & \citet{duan2024sar} \\
      & SD & Semantic Density
      & $\checkmark$ & $\checkmark$ & \citet{qiu2024semantic} \\
      & Eig & Sum of Eigenvalues
      & $\checkmark$ & $\times$ & \citet{lin2024generating} \\
      & Deg & Degree Matrix
      & $\checkmark$ & $\times$ & \citet{lin2024generating} \\
      & Ecc & Eccentricity
      & $\checkmark$ & $\times$ & \citet{lin2024generating} \\
      \bottomrule
  \end{tabular}
  \caption{Overview of the UQ contenders for LLMs.}
  \label{tab:llm_uq}
\end{table}

To adapt these methods to multi-step forecasting, we treat each forecast trajectory as the input of the method, while replacing the token-level distribution with the time-step-level distribution constructed in Subsection~\ref{subsec:sga}. Table~\ref{tab:llm_huq_correspondence} shows the correspondence between the key elements used in UQ for LLMs and HUQ for TSFMs.

\begin{table}[ht]
    \centering
    \footnotesize
    \begin{tabular}{ll}
        \toprule
        UQ for LLMs & HUQ for TSFMs \\
        \midrule
        Token sequence & Forecast trajectory \\
        Token & Forecast value \\
        Token-level distribution & Time-step-level distribution \\
        Semantic equivalence & DTW-based equivalence \\
        Semantic similarity & DTW-based similarity \\
        \bottomrule
    \end{tabular}
    \caption{Correspondence between the key elements used in UQ for LLMs and HUQ for TSFMs.}
    \label{tab:llm_huq_correspondence}
\end{table}

For information-based methods, we calculate the information-theoretic measures directly after modeling forecast trajectories as sequences. 
For diversity-based methods, we additionally replace their text-based measures of semantic equivalence or similarity with a DTW-based criterion when measuring the diversity. Specifically, let $\hat{\boldsymbol{x}}$ and $\hat{\boldsymbol{x}}'$ be two forecast trajectories, and $d_{\text{DTW}}(\hat{\boldsymbol{x}}, \hat{\boldsymbol{x}}')$ denotes the DTW distance between them. Their DTW-based equivalence $\mathcal{E}(\hat{\boldsymbol{x}}, \hat{\boldsymbol{x}}')$ holds when their DTW distance does not exceed the threshold $\tau$, i.e., $\mathcal{E}(\hat{\boldsymbol{x}}, \hat{\boldsymbol{x}}') = \mathbb{I}\{d_{\text{DTW}}(\hat{\boldsymbol{x}}, \hat{\boldsymbol{x}}') \leq \tau\}$. Their DTW-based similarity $s(\hat{\boldsymbol{x}},\hat{\boldsymbol{x}}')$ is defined as $s(\hat{\boldsymbol{x}},\hat{\boldsymbol{x}}') = \exp(-d_{\text{DTW}}(\hat{\boldsymbol{x}},\hat{\boldsymbol{x}}'))$.
The further adaptations are listed as follows.

\begin{itemize}[leftmargin=*]
    \item \textbf{SE.} We replace semantic equivalence between sentences with DTW-based equivalence $\mathcal{E}(\hat{\boldsymbol{x}}, \hat{\boldsymbol{x}}')$ between forecast trajectories.

    \item \textbf{SAR.} We replace token-level and sentence-level relevance with DTW-based forecast-value-level and forecast-trajectory-level relevance, respectively. The forecast-value-level relevance of the value $\hat{x}'$ in the $i$-th trajectory $\hat{\boldsymbol{x}}^i$ is defined as $r^i(\hat{x}') = 1 - s(\hat{\boldsymbol{x}}^i,\hat{\boldsymbol{x}}^i\setminus \hat{x}')$, where $\hat{\boldsymbol{x}}^i\setminus \hat{x}'$ denotes the trajectory with the forecast value $\hat{x}'$ removed. Then we compute the forecast-trajectory-level relevance with DTW-based similarity $\sum_{j=1, j\neq i}^{K}s(\hat{\boldsymbol{x}}^i,\hat{\boldsymbol{x}}^j)p(\hat{\boldsymbol{x}}^j)$, where $p(\hat{\boldsymbol{x}}^j)$ is the forecast-value-level SAR probability of $\hat{\boldsymbol{x}}^j$ and is derived from forecast-value-level relevance.

    \item \textbf{SD.} We replace the semantic similarity kernel between sentences with a DTW-based similarity kernel between forecast trajectories, which is defined as $\mathcal{K}_{\mathrm{DTW}}(\hat{\boldsymbol{x}},\hat{\boldsymbol{x}}') = s(\hat{\boldsymbol{x}},\hat{\boldsymbol{x}}')$.

    \item \textbf{Eig/Deg/Ecc.} We replace the similarity matrix over text sequences with the DTW-based similarity matrix over forecast trajectories. Specifically, the $(i, j)$-th entry $w_{ij}$ of the similarity matrix $\mathbf{W}$ is defined as $s(\hat{\boldsymbol{x}}^i,\hat{\boldsymbol{x}}^j)$, representing the DTW-based similarity between two sampled forecast trajectories $\hat{\boldsymbol{x}}^i$ and $\hat{\boldsymbol{x}}^j$.
\end{itemize}

\subsection{UQ Performance Evaluations} \label{app:performance}
This subsection provides additional details on the evaluation of UQ performance.

\begin{figure}[ht]
    \centering
    \includegraphics[width=0.8\linewidth]{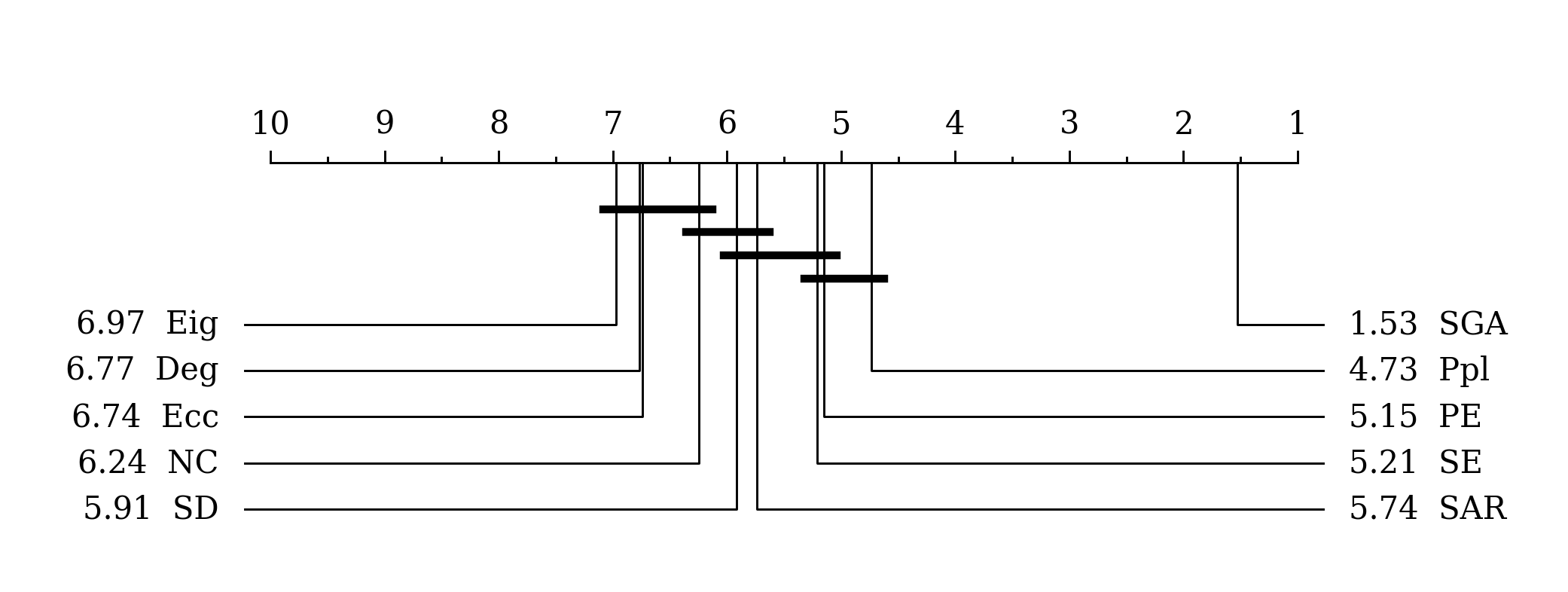}
    \caption{Overall evaluation ranking of 10 UQ methods across 27 datasets and 11 TSFMs.}
    \label{fig:ranking}
\end{figure}

Figure~\ref{fig:ranking} visualizes the overall evaluation ranks of the UQ methods, where the ranking scores are calculated from the average performance across 27 datasets and 11 models. A lower rank indicates better overall performance of a UQ method. It can be observed that SGA achieves the best rank among all methods, further demonstrating the strong advantage of SGA. 

Tables~\ref{tab:all_comparisons_1}--\ref{tab:all_comparisons_11} list the comparisons of NEAURC of SGA and its contenders across 27 datasets for each individual TSFM, where bold and underlined values denote the best and second-best results, respectively. It is observed that SGA exceeds all contenders in 260 dataset-model pairs among a total of $11 \times 27 = 297$ pairs, thus validating the effectiveness and generality of SGA in various kinds of datasets.

\input{Tables/all_comparisons}

\clearpage
\subsection{Relation between Uncertainty and Scale} \label{app:relation}
\begin{figure}[ht]
    \centering
    \includegraphics[width=\linewidth]{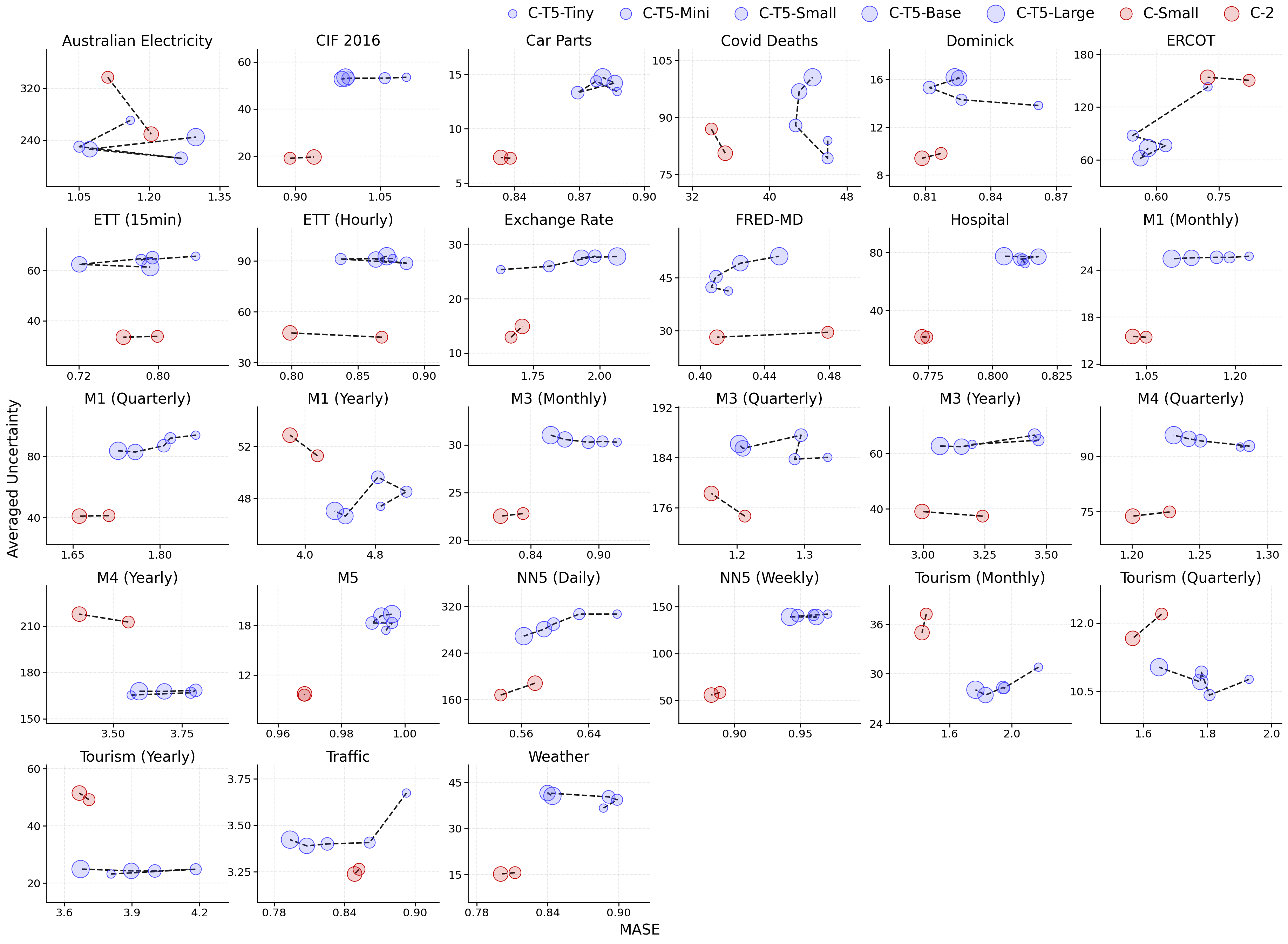}
    \caption{Plots of averaged uncertainty versus MASE of TSFMs over diverse scales across 27 datasets.}
    \label{fig:all_scaling}
\end{figure}

Figure~\ref{fig:all_scaling} shows the plots of averaged uncertainty versus MASE of TSFMs over diverse scales on 27 datasets. 
It is observed that across some individual datasets, larger circles do not appear in the bottom-left direction relative to smaller circles, indicating that the scaling law of TSFM performance or uncertainty is not consistently observable with every single dataset.

\subsection{Additional Ablation Analyses} \label{app:ablation}
\begin{figure}[ht]
    \centering
    \includegraphics[width=\linewidth]{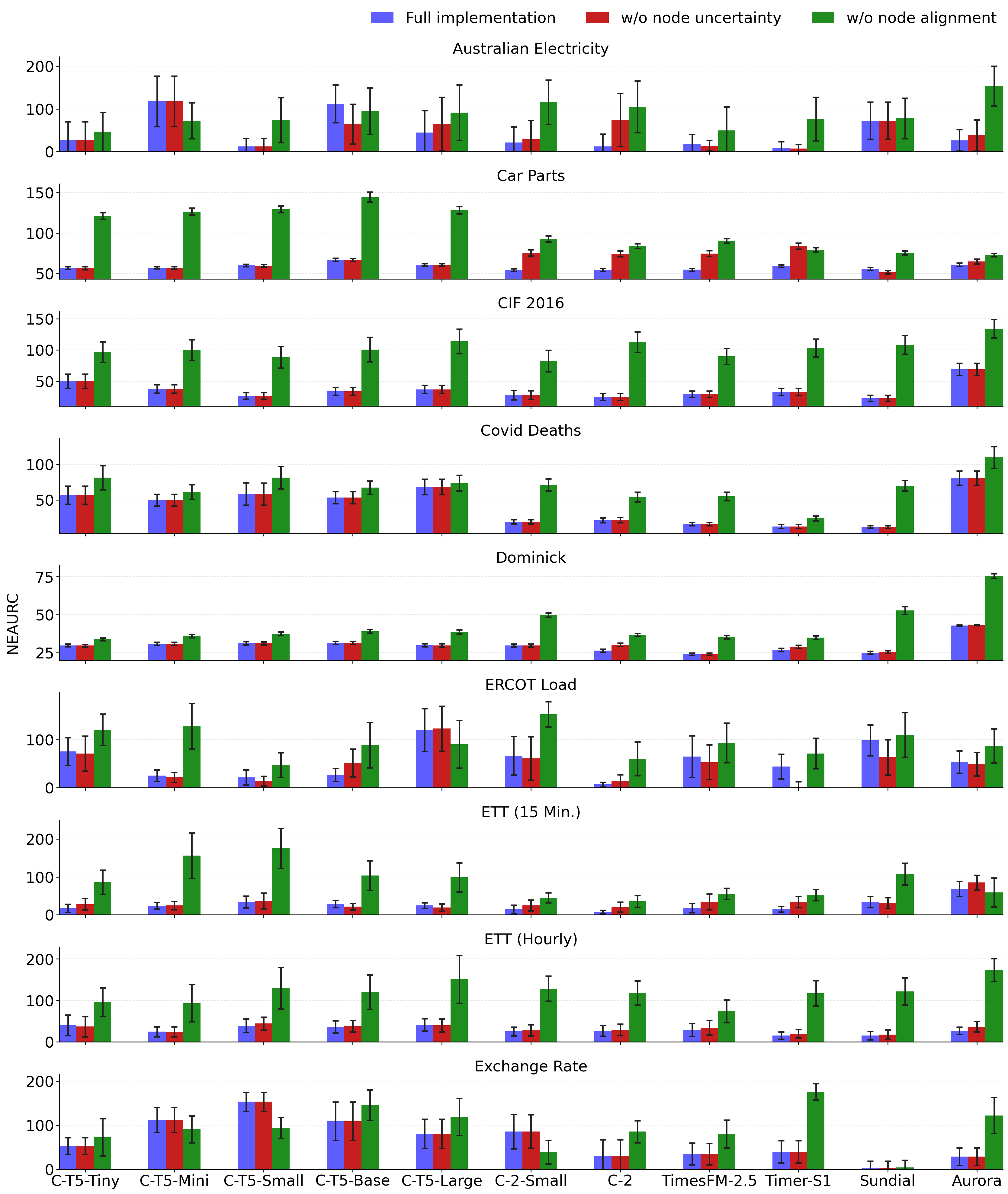}
    \caption{Ablation comparison of UQ performance of SGA on 11 TSFMs of 3 types over the first 9 datasets.}
    \label{fig:all_ablation_1}
\end{figure}

\begin{figure}[ht]
    \centering
    \includegraphics[width=\linewidth]{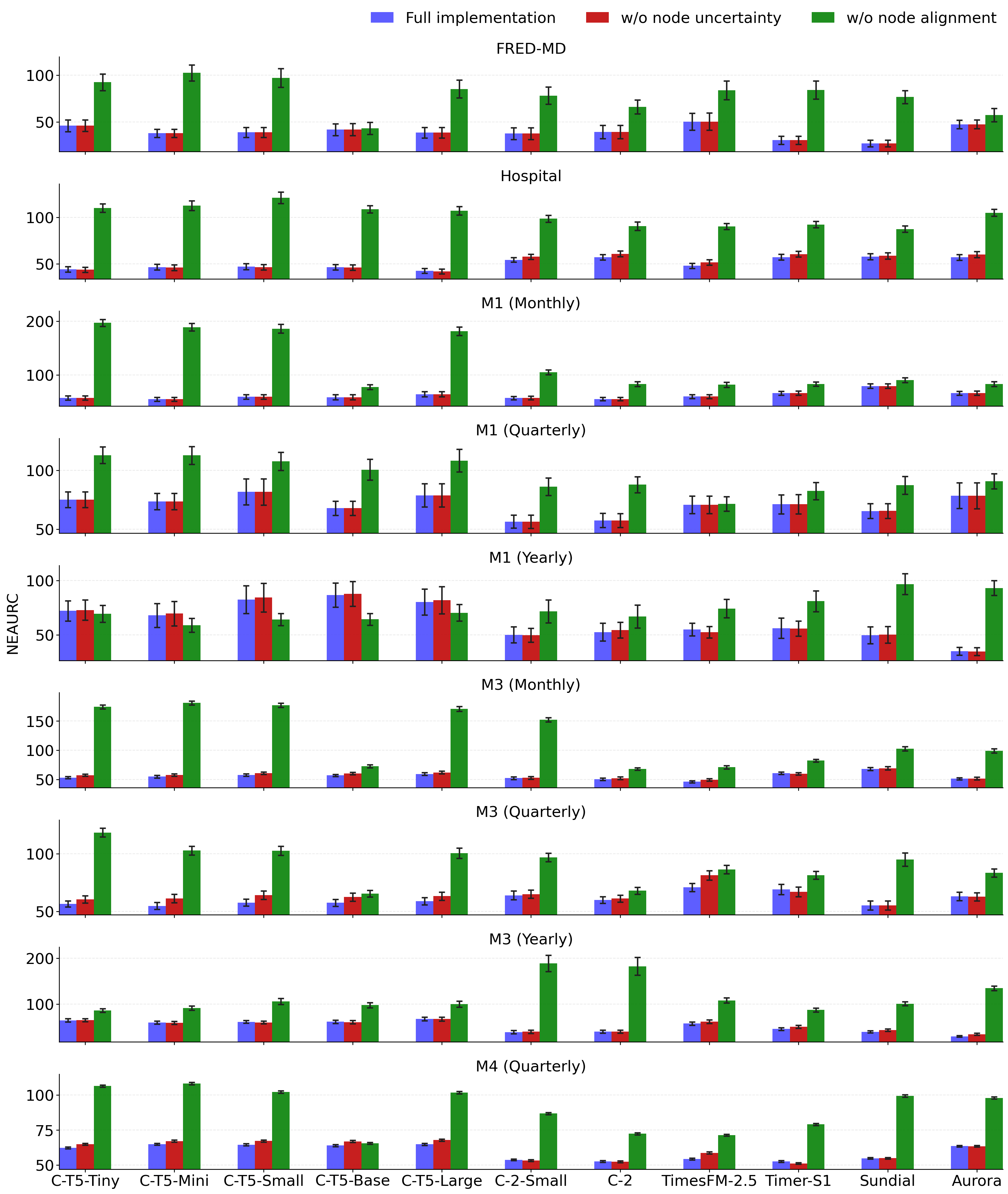}
    \caption{Ablation comparison of UQ performance of SGA on 11 TSFMs of 3 types over the middle 9 datasets.}
    \label{fig:all_ablation_2}
\end{figure}

\begin{figure}[ht]
    \centering
    \includegraphics[width=\linewidth]{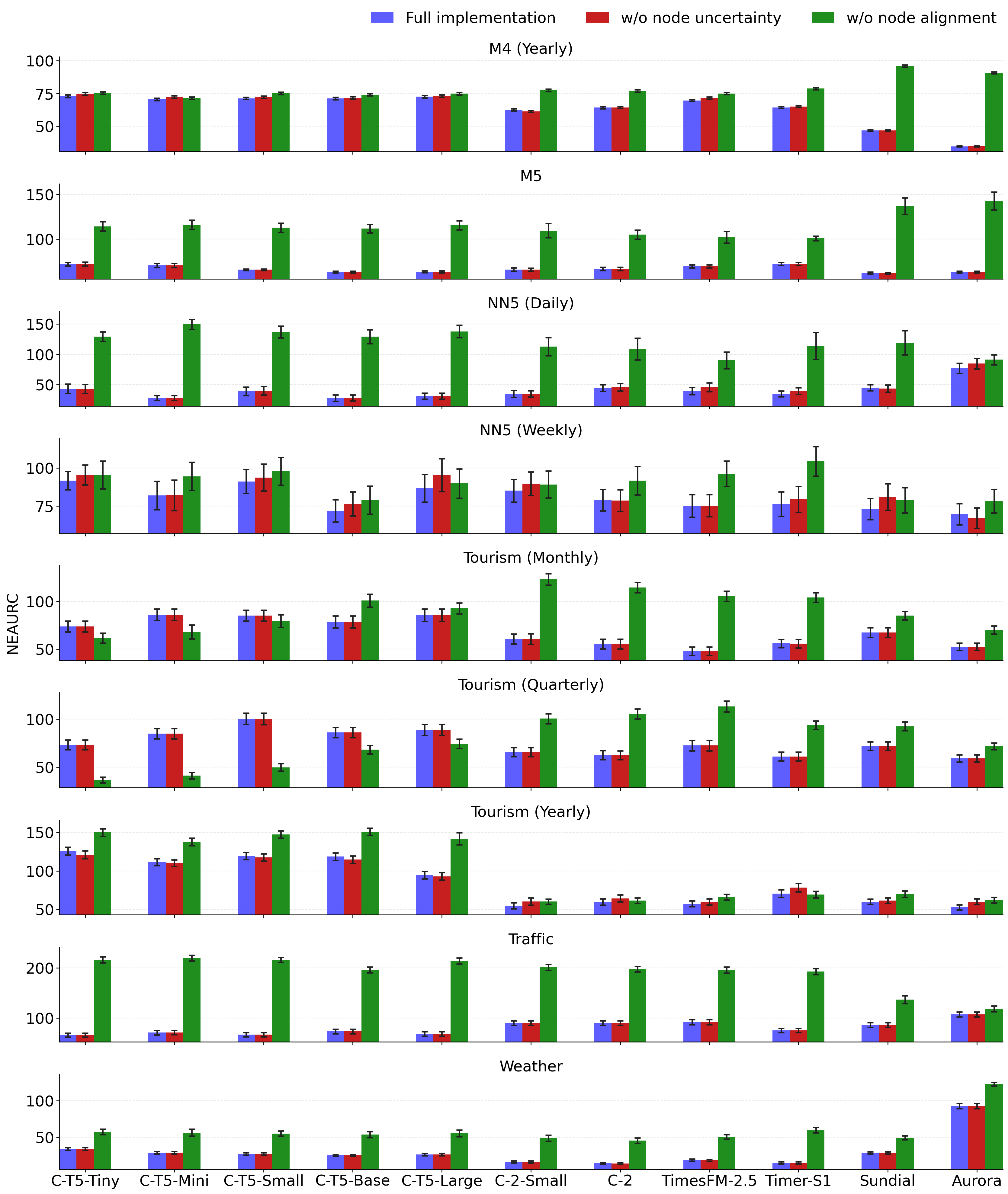}
    \caption{Ablation comparison of UQ performance of SGA on 11 TSFMs of 3 types over the last 9 datasets.}
    \label{fig:all_ablation_3}
\end{figure}

This subsection further analyzes how node alignment and node uncertainty calculation affect the performance of SGA on diverse datasets. Figures~\ref{fig:all_ablation_1},~\ref{fig:all_ablation_2},~\ref{fig:all_ablation_3} show the ablation comparison of UQ performance of SGA on 11 TSFMs of 3 types over 27 datasets.  
There are two key observations. First, the green bars are significantly taller than the blue bars, demonstrating that node alignment is critical to SGA’s performance. Second, the red bars are typically slightly taller than their blue counterparts, which suggests that when applying SGA to selective prediction, topological information contributes more significantly to the construction of graph complexity than TSFM-inherent stochasticity.

\clearpage
\subsection{Additional Sensitivity Analyses} \label{app:sensitivity}
\begin{figure}[ht]
    \centering
    \includegraphics[width=\linewidth]{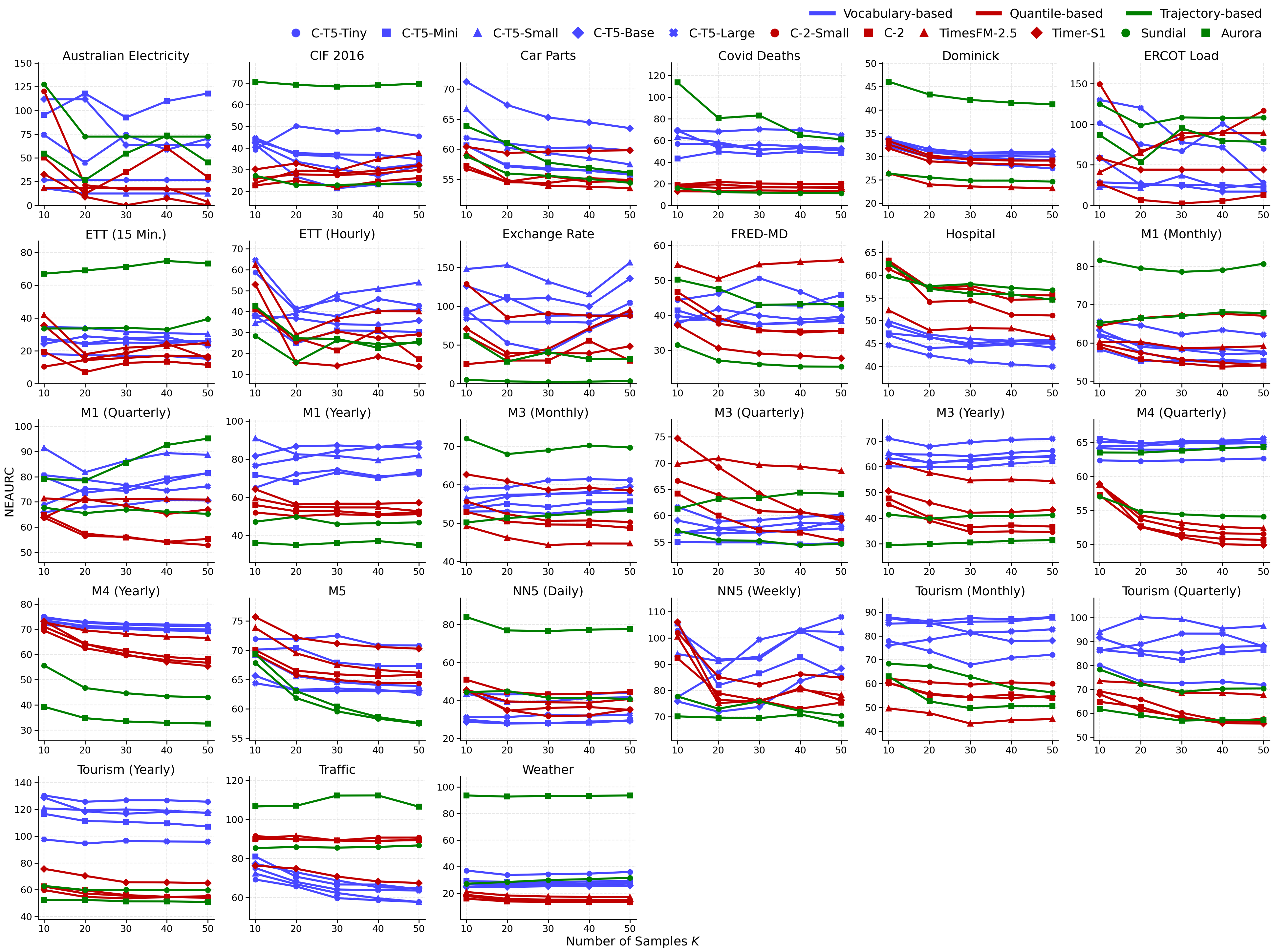}
    \caption{Impact of the number of samples $K$ on the performance of SGA across 27 datasets and 11 TSFMs of 3 types.}
    \label{fig:all_k}
\end{figure}

This subsection further investigates how three key hyperparameters in SGA, namely the number of samples $K$, the slicing length $l_s$, and the threshold coefficient $\lambda$, affect the HUQ performance of SGA on 27 datasets. 
Figure~\ref{fig:all_k} shows the impact of the number of samples $K$ on the performance of SGA across 27 datasets and 11 TSFMs of 3 types. Since a larger $K$ leads to higher computational costs, we recommend $K=20$ for the C-2 family, TimesFM-2.5, Sundial, and Aurora, and $K=30$ for the C-T5 family and Timer-S1 to achieve a favorable trade-off between efficiency and performance.

\begin{figure}[ht]
    \centering
    \includegraphics[width=\linewidth]{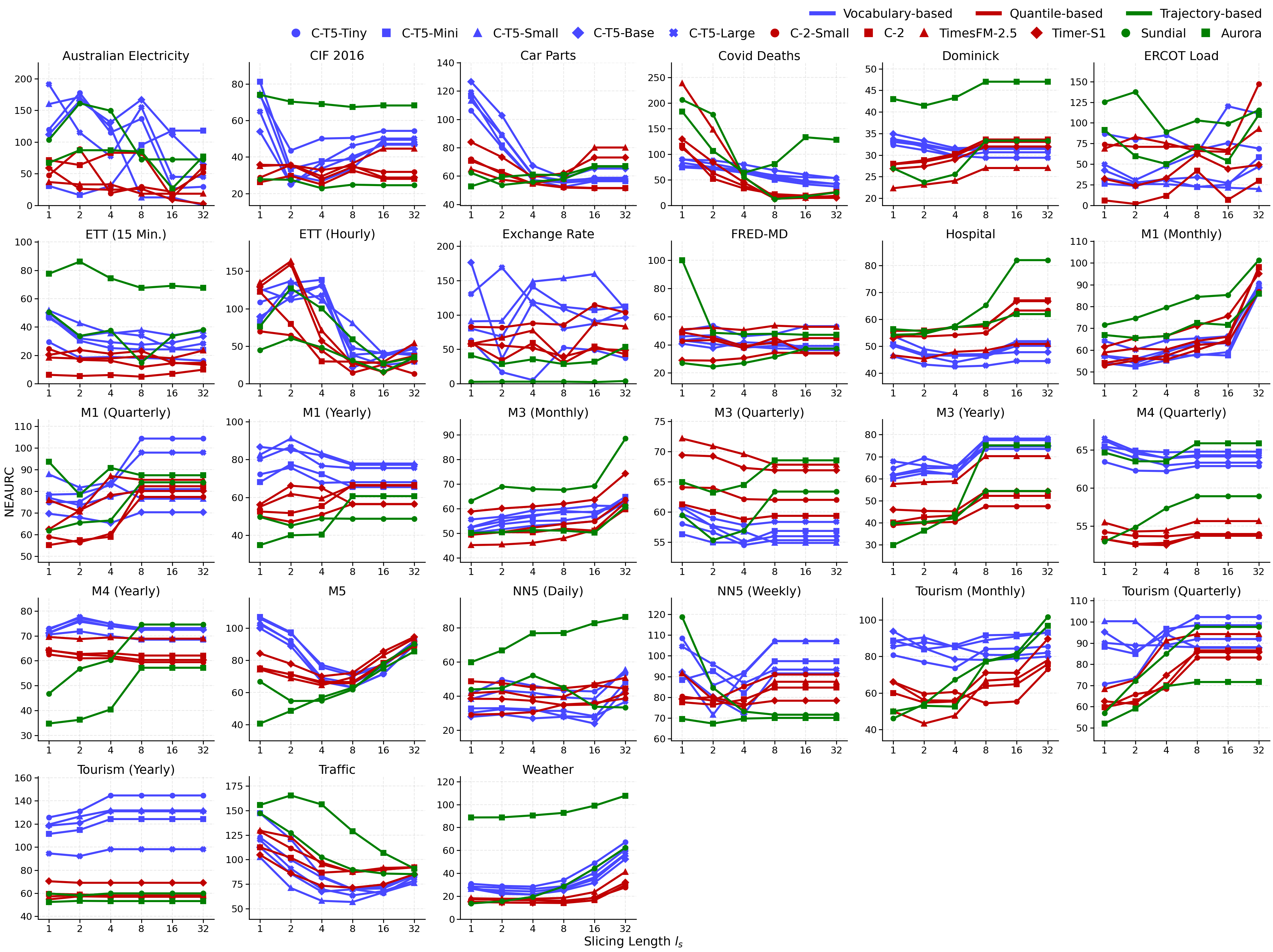}
    \caption{Impact of the slicing length $l_s$ on the performance of SGA across 27 datasets and 11 TSFMs of 3 types.}
    \label{fig:all_l_s}
\end{figure}

\clearpage
Figure~\ref{fig:all_l_s} shows the impact of the slicing length $l_s$ on the performance of SGA across 27 datasets and 11 TSFMs of 3 types. According to the overall performance, we recommend $l_s = 4$.

\begin{figure}[ht]
    \centering
    \includegraphics[width=\linewidth]{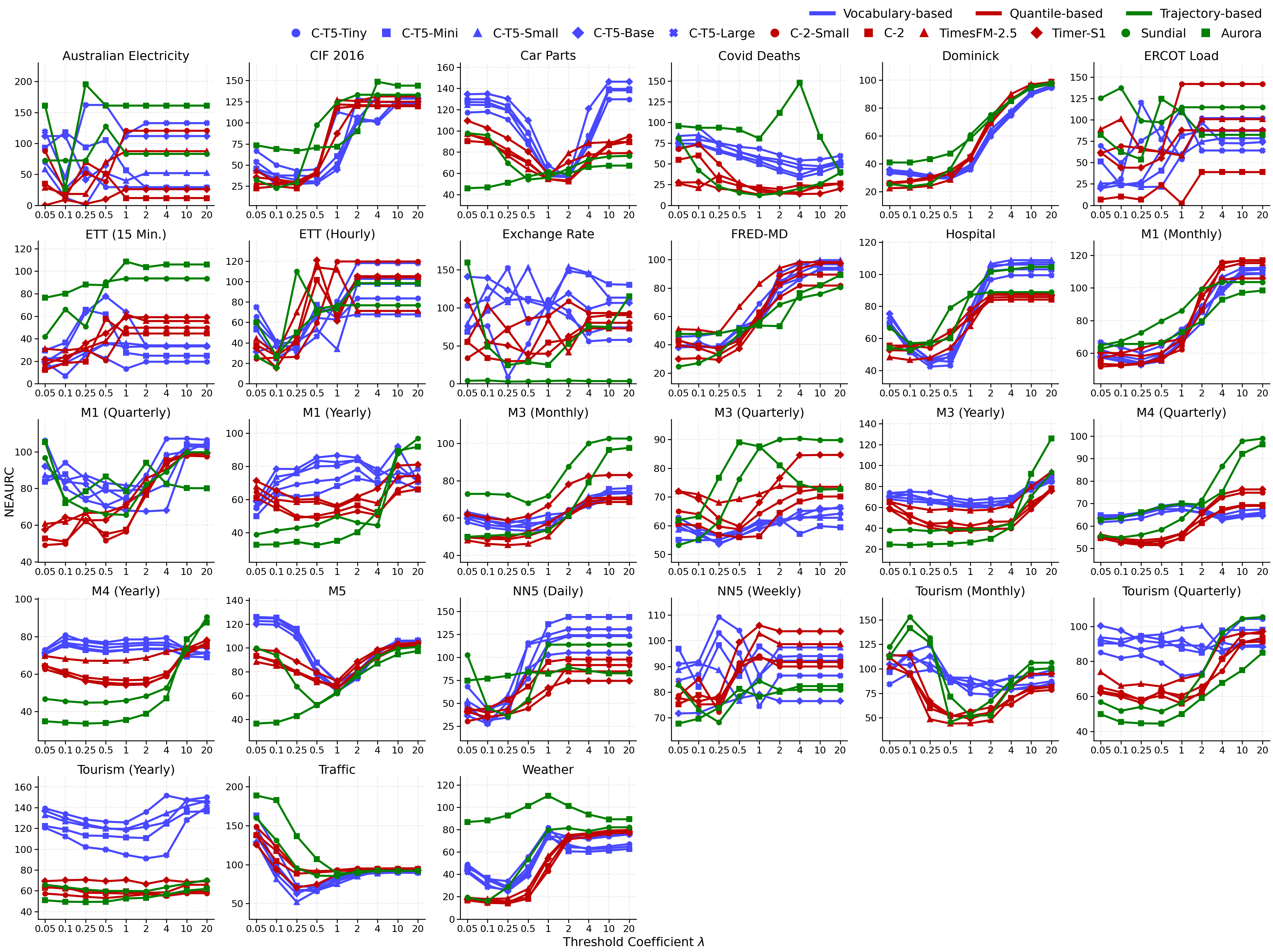}
    \caption{Impact of the threshold coefficient $\lambda$ on the performance of SGA across 27 datasets and 11 TSFMs of 3 types.}
    \label{fig:all_lambda}
\end{figure}

\clearpage
Figure~\ref{fig:all_lambda} shows the impact of the threshold coefficient $\lambda$ on the performance of SGA across 27 datasets and 11 TSFMs of 3 types. According to the overall performance, we recommend $\lambda = 0.25$.

\bibliographystyle{apalike}
\bibliography{SGA}

\end{document}

%% file: Tables/all_comparisons.tex

\begin{table*}[ht]
\centering
\footnotesize
\setlength{\tabcolsep}{1pt}
\renewcommand{\arraystretch}{0.78}
\resizebox{\linewidth}{!}{
\begin{tabular}{lcccccccccc}
\toprule
Dataset & NC & Ppl & PE & Eig & Ecc & Deg & SD & SAR & SE & SGA \\
\midrule
Australian Electricity & \phantom{0}88.70\textsubscript{52.20} & \phantom{0}91.40\textsubscript{62.00} & \underline{\phantom{0}77.40\textsubscript{51.60}} & 121.70\textsubscript{51.00} & 107.00\textsubscript{41.90} & 121.70\textsubscript{50.50} & \phantom{0}91.40\textsubscript{61.50} & 105.10\textsubscript{52.50} & \phantom{0}77.40\textsubscript{49.90} & \textbf{\phantom{0}26.80\textsubscript{43.30}} \\
CIF 2016 & 138.30\textsubscript{21.80} & 176.40\textsubscript{38.10} & 136.50\textsubscript{30.10} & 132.40\textsubscript{17.50} & \underline{126.70\textsubscript{20.70}} & 132.50\textsubscript{17.80} & 169.80\textsubscript{37.10} & 128.90\textsubscript{19.20} & 135.30\textsubscript{30.20} & \textbf{\phantom{0}50.10\textsubscript{11.40}} \\
Car Parts & 118.00\textsubscript{4.40\phantom{0}} & \underline{101.80\textsubscript{3.70\phantom{0}}} & 121.20\textsubscript{4.60\phantom{0}} & 137.60\textsubscript{6.00\phantom{0}} & 142.20\textsubscript{6.30\phantom{0}} & 137.70\textsubscript{5.70\phantom{0}} & 129.50\textsubscript{5.80\phantom{0}} & 123.70\textsubscript{4.50\phantom{0}} & 121.20\textsubscript{4.20\phantom{0}} & \textbf{\phantom{0}57.10\textsubscript{1.70\phantom{0}}} \\
Covid Deaths & \phantom{0}76.20\textsubscript{13.30} & \underline{\phantom{0}59.10\textsubscript{10.70}} & \phantom{0}79.80\textsubscript{17.90} & \phantom{0}89.70\textsubscript{15.70} & \phantom{0}95.60\textsubscript{15.60} & \phantom{0}87.60\textsubscript{16.30} & \phantom{0}72.20\textsubscript{8.00\phantom{0}} & \phantom{0}93.00\textsubscript{17.00} & \phantom{0}83.70\textsubscript{17.00} & \textbf{\phantom{0}56.70\textsubscript{12.70}} \\
\bottomrule
\end{tabular}}
\caption{Comparisons of NEAURC of SGA and its contenders across 27 datasets for Chronos-T5-Tiny, where bold and underlined values denote the best and second-best results, respectively.}
\label{tab:all_comparisons_1}
\end{table*}

\begin{table*}[ht]
\centering
\footnotesize
\setlength{\tabcolsep}{1pt}
\renewcommand{\arraystretch}{0.78}
\resizebox{\linewidth}{!}{
\begin{tabular}{lcccccccccc}
\toprule
Dataset & NC & Ppl & PE & Eig & Ecc & Deg & SD & SAR & SE & SGA \\
\midrule
Australian Electricity & 182.70\textsubscript{38.60} & 106.50\textsubscript{49.40} & \textbf{\phantom{0}9.10\textsubscript{11.70}} & 187.40\textsubscript{27.30} & \phantom{0}76.70\textsubscript{43.90} & 187.40\textsubscript{20.80} & 106.50\textsubscript{69.00} & 187.40\textsubscript{25.30} & \underline{\phantom{0}9.10\textsubscript{11.00}} & 118.00\textsubscript{59.10} \\
CIF 2016 & 131.80\textsubscript{16.30} & 111.40\textsubscript{15.00} & 128.20\textsubscript{26.50} & 119.60\textsubscript{15.40} & \underline{100.80\textsubscript{10.20}} & 119.70\textsubscript{16.00} & 111.20\textsubscript{13.90} & 121.40\textsubscript{16.40} & 129.90\textsubscript{27.80} & \textbf{\phantom{0}37.70\textsubscript{6.60\phantom{0}}} \\
Car Parts & 120.10\textsubscript{4.30\phantom{0}} & \underline{114.40\textsubscript{3.90\phantom{0}}} & 130.20\textsubscript{5.00\phantom{0}} & 146.70\textsubscript{6.30\phantom{0}} & 139.40\textsubscript{6.50\phantom{0}} & 146.40\textsubscript{6.30\phantom{0}} & 137.40\textsubscript{6.00\phantom{0}} & 131.90\textsubscript{5.20\phantom{0}} & 127.40\textsubscript{4.20\phantom{0}} & \textbf{\phantom{0}57.30\textsubscript{1.40\phantom{0}}} \\
Covid Deaths & \phantom{0}87.30\textsubscript{12.90} & \underline{\phantom{0}59.70\textsubscript{11.50}} & \phantom{0}69.30\textsubscript{13.70} & \phantom{0}81.70\textsubscript{12.80} & \phantom{0}81.60\textsubscript{12.80} & \phantom{0}81.00\textsubscript{12.90} & \phantom{0}60.20\textsubscript{36.80} & \phantom{0}77.60\textsubscript{13.20} & \phantom{0}75.60\textsubscript{14.30} & \textbf{\phantom{0}49.70\textsubscript{8.40\phantom{0}}} \\
\bottomrule
\end{tabular}}
\caption{Comparisons of NEAURC of SGA and its contenders across 27 datasets for Chronos-T5-Mini, where bold and underlined values denote the best and second-best results, respectively.}
\label{tab:all_comparisons_2}
\end{table*}

\begin{table*}[ht]
\centering
\footnotesize
\setlength{\tabcolsep}{1pt}
\renewcommand{\arraystretch}{0.78}
\resizebox{\linewidth}{!}{
\begin{tabular}{lcccccccccc}
\toprule
Dataset & NC & Ppl & PE & Eig & Ecc & Deg & SD & SAR & SE & SGA \\
\midrule
Australian Electricity & \phantom{0}69.00\textsubscript{41.20} & \phantom{0}71.70\textsubscript{48.60} & \phantom{0}85.40\textsubscript{43.40} & \underline{\phantom{0}52.30\textsubscript{35.90}} & \phantom{0}95.60\textsubscript{46.60} & \phantom{0}52.30\textsubscript{36.60} & \phantom{0}71.70\textsubscript{47.20} & \phantom{0}52.30\textsubscript{36.70} & \phantom{0}85.40\textsubscript{42.30} & \textbf{\phantom{0}12.40\textsubscript{18.80}} \\
CIF 2016 & 114.50\textsubscript{16.40} & 156.90\textsubscript{35.10} & 118.70\textsubscript{28.70} & 121.00\textsubscript{15.90} & \underline{108.90\textsubscript{14.50}} & 121.00\textsubscript{16.40} & 159.20\textsubscript{36.90} & 115.30\textsubscript{16.20} & 117.80\textsubscript{29.10} & \textbf{\phantom{0}26.80\textsubscript{5.30\phantom{0}}} \\
Car Parts & 125.20\textsubscript{5.60\phantom{0}} & \underline{118.00\textsubscript{3.90\phantom{0}}} & 128.50\textsubscript{5.40\phantom{0}} & 139.70\textsubscript{6.30\phantom{0}} & 132.50\textsubscript{6.20\phantom{0}} & 139.50\textsubscript{6.30\phantom{0}} & 134.80\textsubscript{6.30\phantom{0}} & 130.30\textsubscript{5.60\phantom{0}} & 126.30\textsubscript{4.60\phantom{0}} & \textbf{\phantom{0}60.20\textsubscript{1.60\phantom{0}}} \\
Covid Deaths & \phantom{0}83.20\textsubscript{12.70} & \phantom{0}73.30\textsubscript{16.30} & \underline{\phantom{0}65.70\textsubscript{11.60}} & \phantom{0}82.10\textsubscript{8.90\phantom{0}} & \phantom{0}80.80\textsubscript{8.90\phantom{0}} & \phantom{0}81.90\textsubscript{9.30\phantom{0}} & \phantom{0}76.20\textsubscript{9.30\phantom{0}} & \phantom{0}82.50\textsubscript{10.80} & \phantom{0}73.30\textsubscript{11.30} & \textbf{\phantom{0}58.30\textsubscript{15.50}} \\
\bottomrule
\end{tabular}}
\caption{Comparisons of NEAURC of SGA and its contenders across 27 datasets for Chronos-T5-Small, where bold and underlined values denote the best and second-best results, respectively.}
\label{tab:all_comparisons_3}
\end{table*}

\begin{table*}[ht]
\centering
\footnotesize
\setlength{\tabcolsep}{1pt}
\renewcommand{\arraystretch}{0.78}
\resizebox{\linewidth}{!}{
\begin{tabular}{lcccccccccc}
\toprule
Dataset & NC & Ppl & PE & Eig & Ecc & Deg & SD & SAR & SE & SGA \\
\midrule
Australian Electricity & \phantom{0}89.60\textsubscript{54.50} & 118.40\textsubscript{54.30} & 119.70\textsubscript{42.10} & \textbf{\phantom{0}68.10\textsubscript{46.70}} & 148.50\textsubscript{46.00} & \underline{\phantom{0}68.10\textsubscript{46.30}} & 118.40\textsubscript{54.20} & \phantom{0}83.20\textsubscript{53.70} & 119.70\textsubscript{42.40} & 111.90\textsubscript{44.10} \\
CIF 2016 & 117.70\textsubscript{16.30} & 109.80\textsubscript{26.20} & 101.00\textsubscript{21.20} & 109.00\textsubscript{15.60} & \underline{\phantom{0}93.00\textsubscript{13.50}} & 109.10\textsubscript{16.00} & 111.10\textsubscript{27.30} & 103.80\textsubscript{16.90} & 101.50\textsubscript{21.60} & \textbf{\phantom{0}33.70\textsubscript{6.30\phantom{0}}} \\
Car Parts & 132.80\textsubscript{5.10\phantom{0}} & \underline{119.30\textsubscript{3.70\phantom{0}}} & 130.20\textsubscript{5.00\phantom{0}} & 141.30\textsubscript{6.70\phantom{0}} & 133.30\textsubscript{6.80\phantom{0}} & 141.70\textsubscript{6.70\phantom{0}} & 140.90\textsubscript{6.40\phantom{0}} & 132.80\textsubscript{5.50\phantom{0}} & 132.90\textsubscript{5.10\phantom{0}} & \textbf{\phantom{0}67.30\textsubscript{1.90\phantom{0}}} \\
Covid Deaths & \phantom{0}91.90\textsubscript{10.00} & \underline{\phantom{0}64.00\textsubscript{10.60}} & \phantom{0}75.10\textsubscript{10.50} & \phantom{0}97.70\textsubscript{11.00} & \phantom{0}96.40\textsubscript{12.20} & \phantom{0}95.10\textsubscript{10.80} & \phantom{0}69.30\textsubscript{25.00} & \phantom{0}86.40\textsubscript{11.20} & \phantom{0}79.70\textsubscript{10.10} & \textbf{\phantom{0}53.20\textsubscript{8.50\phantom{0}}} \\
\bottomrule
\end{tabular}}
\caption{Comparisons of NEAURC of SGA and its contenders across 27 datasets for Chronos-T5-Base, where bold and underlined values denote the best and second-best results, respectively.}
\label{tab:all_comparisons_4}
\end{table*}

\begin{table*}[ht]
\centering
\footnotesize
\setlength{\tabcolsep}{1pt}
\renewcommand{\arraystretch}{0.78}
\resizebox{\linewidth}{!}{
\begin{tabular}{lcccccccccc}
\toprule
Dataset & NC & Ppl & PE & Eig & Ecc & Deg & SD & SAR & SE & SGA \\
\midrule
Australian Electricity & 100.50\textsubscript{45.70} & 142.80\textsubscript{44.10} & 116.90\textsubscript{56.10} & \underline{\phantom{0}85.10\textsubscript{47.20}} & \phantom{0}89.60\textsubscript{48.50} & \phantom{0}85.10\textsubscript{47.20} & 142.80\textsubscript{42.70} & \phantom{0}85.10\textsubscript{46.00} & 116.90\textsubscript{55.90} & \textbf{\phantom{0}44.90\textsubscript{51.60}} \\
CIF 2016 & 121.30\textsubscript{17.10} & 176.10\textsubscript{36.40} & 115.30\textsubscript{24.70} & 107.60\textsubscript{15.00} & 136.30\textsubscript{21.10} & 107.60\textsubscript{15.40} & 176.40\textsubscript{37.90} & \underline{103.90\textsubscript{15.90}} & 114.90\textsubscript{26.20} & \textbf{\phantom{0}36.90\textsubscript{6.60\phantom{0}}} \\
Car Parts & 126.00\textsubscript{5.30\phantom{0}} & \underline{113.50\textsubscript{3.60\phantom{0}}} & 130.40\textsubscript{4.70\phantom{0}} & 142.40\textsubscript{6.40\phantom{0}} & 140.10\textsubscript{7.00\phantom{0}} & 142.60\textsubscript{6.50\phantom{0}} & 136.50\textsubscript{5.90\phantom{0}} & 132.00\textsubscript{4.70\phantom{0}} & 126.90\textsubscript{4.10\phantom{0}} & \textbf{\phantom{0}60.90\textsubscript{1.50\phantom{0}}} \\
Covid Deaths & 102.80\textsubscript{12.70} & \underline{\phantom{0}81.10\textsubscript{16.60}} & \phantom{0}93.50\textsubscript{17.40} & 102.30\textsubscript{14.30} & 100.60\textsubscript{13.20} & 102.90\textsubscript{14.90} & \phantom{0}81.40\textsubscript{19.50} & \phantom{0}99.30\textsubscript{15.30} & \phantom{0}87.10\textsubscript{13.70} & \textbf{\phantom{0}68.20\textsubscript{10.70}} \\
\bottomrule
\end{tabular}}
\caption{Comparisons of NEAURC of SGA and its contenders across 27 datasets for Chronos-T5-Large, where bold and underlined values denote the best and second-best results, respectively.}
\label{tab:all_comparisons_5}
\end{table*}

\begin{table*}[ht]
\centering
\footnotesize
\setlength{\tabcolsep}{1pt}
\renewcommand{\arraystretch}{0.78}
\resizebox{\linewidth}{!}{
\begin{tabular}{lcccccccccc}
\toprule
Dataset & NC & Ppl & PE & Eig & Ecc & Deg & SD & SAR & SE & SGA \\
\midrule
Australian Electricity & \phantom{0}63.40\textsubscript{42.30} & \phantom{0}63.40\textsubscript{44.60} & 120.40\textsubscript{43.20} & 120.40\textsubscript{42.40} & 129.80\textsubscript{43.60} & 120.40\textsubscript{41.30} & \underline{\phantom{0}52.00\textsubscript{38.30}} & 120.40\textsubscript{42.80} & 120.40\textsubscript{43.50} & \textbf{\phantom{0}21.20\textsubscript{36.90}} \\
CIF 2016 & 118.40\textsubscript{16.50} & 112.60\textsubscript{15.50} & 118.40\textsubscript{16.50} & 112.60\textsubscript{15.20} & 122.50\textsubscript{18.30} & 114.70\textsubscript{15.80} & \underline{108.60\textsubscript{14.60}} & 113.70\textsubscript{15.60} & 116.80\textsubscript{15.90} & \textbf{\phantom{0}27.90\textsubscript{7.40\phantom{0}}} \\
Car Parts & 129.90\textsubscript{6.90\phantom{0}} & 127.20\textsubscript{5.60\phantom{0}} & 136.10\textsubscript{5.40\phantom{0}} & 131.20\textsubscript{6.20\phantom{0}} & \underline{121.80\textsubscript{5.40\phantom{0}}} & 131.80\textsubscript{6.40\phantom{0}} & 136.50\textsubscript{6.40\phantom{0}} & 123.80\textsubscript{4.60\phantom{0}} & 122.40\textsubscript{4.40\phantom{0}} & \textbf{\phantom{0}54.50\textsubscript{1.70\phantom{0}}} \\
Covid Deaths & \phantom{0}69.30\textsubscript{9.10\phantom{0}} & \underline{\phantom{0}56.00\textsubscript{7.30\phantom{0}}} & \phantom{0}71.60\textsubscript{9.90\phantom{0}} & \phantom{0}63.70\textsubscript{7.50\phantom{0}} & \phantom{0}63.70\textsubscript{7.30\phantom{0}} & \phantom{0}63.90\textsubscript{7.30\phantom{0}} & \phantom{0}62.70\textsubscript{7.20\phantom{0}} & \phantom{0}67.90\textsubscript{9.30\phantom{0}} & \phantom{0}70.30\textsubscript{9.90\phantom{0}} & \textbf{\phantom{0}19.40\textsubscript{3.00\phantom{0}}} \\
\bottomrule
\end{tabular}}
\caption{Comparisons of NEAURC of SGA and its contenders across 27 datasets for Chronos-2-Small, where bold and underlined values denote the best and second-best results, respectively.}
\label{tab:all_comparisons_6}
\end{table*}

\begin{table*}[ht]
\centering
\footnotesize
\setlength{\tabcolsep}{1pt}
\renewcommand{\arraystretch}{0.78}
\resizebox{\linewidth}{!}{
\begin{tabular}{lcccccccccc}
\toprule
Dataset & NC & Ppl & PE & Eig & Ecc & Deg & SD & SAR & SE & SGA \\
\midrule
Australian Electricity & \phantom{0}66.00\textsubscript{54.90} & \underline{\phantom{0}43.40\textsubscript{36.10}} & \phantom{0}65.50\textsubscript{52.30} & \phantom{0}89.30\textsubscript{58.10} & 138.90\textsubscript{47.80} & \phantom{0}89.30\textsubscript{57.80} & \phantom{0}43.40\textsubscript{34.40} & \phantom{0}89.30\textsubscript{58.40} & \phantom{0}65.50\textsubscript{51.70} & \textbf{\phantom{0}12.20\textsubscript{28.80}} \\
CIF 2016 & 111.50\textsubscript{15.70} & 100.60\textsubscript{15.90} & 105.40\textsubscript{14.50} & 106.60\textsubscript{14.80} & 113.40\textsubscript{16.00} & 106.70\textsubscript{15.00} & \underline{\phantom{0}94.20\textsubscript{13.80}} & 106.60\textsubscript{15.20} & 105.20\textsubscript{14.70} & \textbf{\phantom{0}24.90\textsubscript{5.40\phantom{0}}} \\
Car Parts & 117.90\textsubscript{5.60\phantom{0}} & \underline{109.90\textsubscript{4.00\phantom{0}}} & 130.90\textsubscript{5.30\phantom{0}} & 123.20\textsubscript{5.50\phantom{0}} & 114.20\textsubscript{4.60\phantom{0}} & 123.70\textsubscript{5.60\phantom{0}} & 124.50\textsubscript{4.30\phantom{0}} & 116.50\textsubscript{4.40\phantom{0}} & 120.70\textsubscript{4.20\phantom{0}} & \textbf{\phantom{0}54.60\textsubscript{1.70\phantom{0}}} \\
Covid Deaths & \phantom{0}70.00\textsubscript{8.50\phantom{0}} & \phantom{0}54.20\textsubscript{9.40\phantom{0}} & \phantom{0}55.90\textsubscript{8.60\phantom{0}} & \phantom{0}63.70\textsubscript{7.00\phantom{0}} & \phantom{0}62.80\textsubscript{7.00\phantom{0}} & \phantom{0}63.80\textsubscript{6.80\phantom{0}} & \phantom{0}63.60\textsubscript{8.00\phantom{0}} & \phantom{0}54.50\textsubscript{6.80\phantom{0}} & \underline{\phantom{0}51.50\textsubscript{6.00\phantom{0}}} & \textbf{\phantom{0}21.70\textsubscript{3.40\phantom{0}}} \\
\bottomrule
\end{tabular}}
\caption{Comparisons of NEAURC of SGA and its contenders across 27 datasets for Chronos-2, where bold and underlined values denote the best and second-best results, respectively.}
\label{tab:all_comparisons_7}
\end{table*}

\begin{table*}[ht]
\centering
\footnotesize
\setlength{\tabcolsep}{1pt}
\renewcommand{\arraystretch}{0.78}
\resizebox{\linewidth}{!}{
\begin{tabular}{lcccccccccc}
\toprule
Dataset & NC & Ppl & PE & Eig & Ecc & Deg & SD & SAR & SE & SGA \\
\midrule
Australian Electricity & 115.50\textsubscript{52.80} & \underline{\phantom{0}67.50\textsubscript{63.10}} & \phantom{0}91.20\textsubscript{55.70} & 169.20\textsubscript{39.90} & \phantom{0}84.60\textsubscript{55.60} & 169.20\textsubscript{39.00} & \phantom{0}87.00\textsubscript{57.00} & 115.50\textsubscript{53.70} & \phantom{0}91.20\textsubscript{55.30} & \textbf{\phantom{0}18.30\textsubscript{22.30}} \\
CIF 2016 & 122.90\textsubscript{18.80} & 101.00\textsubscript{15.70} & \underline{\phantom{0}99.50\textsubscript{16.20}} & 123.10\textsubscript{17.50} & 143.30\textsubscript{23.50} & 125.40\textsubscript{19.20} & 101.60\textsubscript{16.80} & 108.60\textsubscript{18.80} & 105.80\textsubscript{20.10} & \textbf{\phantom{0}29.40\textsubscript{5.10\phantom{0}}} \\
Car Parts & 142.50\textsubscript{7.50\phantom{0}} & \phantom{0}97.90\textsubscript{3.10\phantom{0}} & \phantom{0}98.10\textsubscript{3.90\phantom{0}} & 136.70\textsubscript{6.20\phantom{0}} & 133.10\textsubscript{6.60\phantom{0}} & 136.70\textsubscript{6.40\phantom{0}} & 111.40\textsubscript{4.20\phantom{0}} & \phantom{0}87.90\textsubscript{2.90\phantom{0}} & \underline{\phantom{0}86.80\textsubscript{2.80\phantom{0}}} & \textbf{\phantom{0}54.80\textsubscript{1.60\phantom{0}}} \\
Covid Deaths & \phantom{0}71.70\textsubscript{8.10\phantom{0}} & \underline{\phantom{0}37.80\textsubscript{4.70\phantom{0}}} & \phantom{0}53.70\textsubscript{7.00\phantom{0}} & \phantom{0}77.00\textsubscript{9.10\phantom{0}} & \phantom{0}81.70\textsubscript{9.70\phantom{0}} & \phantom{0}77.80\textsubscript{9.20\phantom{0}} & \phantom{0}59.70\textsubscript{5.80\phantom{0}} & \phantom{0}49.80\textsubscript{5.00\phantom{0}} & \phantom{0}59.40\textsubscript{7.90\phantom{0}} & \textbf{\phantom{0}16.20\textsubscript{2.40\phantom{0}}} \\
\bottomrule
\end{tabular}}
\caption{Comparisons of NEAURC of SGA and its contenders across 27 datasets for TimesFM-2.5, where bold and underlined values denote the best and second-best results, respectively.}
\label{tab:all_comparisons_8}
\end{table*}

\begin{table*}[ht]
\centering
\footnotesize
\setlength{\tabcolsep}{1pt}
\renewcommand{\arraystretch}{0.78}
\resizebox{\linewidth}{!}{
\begin{tabular}{lcccccccccc}
\toprule
Dataset & NC & Ppl & PE & Eig & Ecc & Deg & SD & SAR & SE & SGA \\
\midrule
Australian Electricity & \phantom{0}90.00\textsubscript{57.40} & \underline{\phantom{0}81.00\textsubscript{53.90}} & \phantom{0}92.40\textsubscript{57.60} & \phantom{0}92.40\textsubscript{58.20} & 133.80\textsubscript{57.30} & \phantom{0}92.40\textsubscript{58.00} & 107.60\textsubscript{46.30} & \phantom{0}92.40\textsubscript{58.80} & \phantom{0}92.40\textsubscript{57.70} & \textbf{\phantom{0}8.90\textsubscript{14.50}} \\
CIF 2016 & 114.60\textsubscript{18.70} & \underline{\phantom{0}99.70\textsubscript{16.60}} & 102.40\textsubscript{14.10} & 105.10\textsubscript{15.10} & 111.80\textsubscript{16.30} & 105.60\textsubscript{15.10} & 111.40\textsubscript{15.90} & 104.30\textsubscript{14.90} & 102.00\textsubscript{14.00} & \textbf{\phantom{0}32.90\textsubscript{6.00\phantom{0}}} \\
Car Parts & 158.40\textsubscript{7.20\phantom{0}} & 112.70\textsubscript{3.80\phantom{0}} & 127.50\textsubscript{5.50\phantom{0}} & 160.20\textsubscript{7.00\phantom{0}} & 154.30\textsubscript{7.10\phantom{0}} & 160.30\textsubscript{7.10\phantom{0}} & 123.90\textsubscript{3.70\phantom{0}} & \underline{106.70\textsubscript{3.40\phantom{0}}} & 115.50\textsubscript{4.40\phantom{0}} & \textbf{\phantom{0}59.30\textsubscript{1.50\phantom{0}}} \\
Covid Deaths & \phantom{0}67.00\textsubscript{7.10\phantom{0}} & \underline{\phantom{0}25.20\textsubscript{3.40\phantom{0}}} & \phantom{0}25.20\textsubscript{3.70\phantom{0}} & \phantom{0}68.80\textsubscript{7.10\phantom{0}} & \phantom{0}68.00\textsubscript{7.10\phantom{0}} & \phantom{0}68.20\textsubscript{6.80\phantom{0}} & \phantom{0}59.20\textsubscript{5.60\phantom{0}} & \phantom{0}34.70\textsubscript{4.40\phantom{0}} & \phantom{0}31.90\textsubscript{3.80\phantom{0}} & \textbf{\phantom{0}12.80\textsubscript{2.70\phantom{0}}} \\
\bottomrule
\end{tabular}}
\caption{Comparisons of NEAURC of SGA and its contenders across 27 datasets for Timer-S1, where bold and underlined values denote the best and second-best results, respectively.}
\label{tab:all_comparisons_9}
\end{table*}

\begin{table*}[ht]
\centering
\footnotesize
\setlength{\tabcolsep}{1pt}
\renewcommand{\arraystretch}{0.78}
\resizebox{\linewidth}{!}{
\begin{tabular}{lcccccccccc}
\toprule
Dataset & NC & Ppl & PE & Eig & Ecc & Deg & SD & SAR & SE & SGA \\
\midrule
Australian Electricity & 122.00\textsubscript{53.60} & 125.00\textsubscript{51.60} & 125.00\textsubscript{53.00} & \underline{111.50\textsubscript{52.30}} & 132.90\textsubscript{70.40} & 111.50\textsubscript{53.00} & 149.40\textsubscript{47.10} & 125.00\textsubscript{53.10} & 125.00\textsubscript{51.80} & \textbf{\phantom{0}72.40\textsubscript{43.30}} \\
CIF 2016 & 111.10\textsubscript{16.00} & 111.50\textsubscript{14.60} & 110.70\textsubscript{14.10} & 105.70\textsubscript{14.40} & \underline{\phantom{0}85.50\textsubscript{11.10}} & 104.30\textsubscript{14.00} & \phantom{0}99.60\textsubscript{12.60} & 107.50\textsubscript{14.10} & 108.40\textsubscript{13.70} & \textbf{\phantom{0}22.90\textsubscript{4.80\phantom{0}}} \\
Car Parts & 145.50\textsubscript{6.40\phantom{0}} & 146.20\textsubscript{6.00\phantom{0}} & 150.60\textsubscript{5.90\phantom{0}} & 140.60\textsubscript{5.40\phantom{0}} & \underline{127.20\textsubscript{5.60\phantom{0}}} & 140.70\textsubscript{5.50\phantom{0}} & 143.60\textsubscript{5.50\phantom{0}} & 144.20\textsubscript{5.80\phantom{0}} & 151.00\textsubscript{5.70\phantom{0}} & \textbf{\phantom{0}55.90\textsubscript{1.60\phantom{0}}} \\
Covid Deaths & \phantom{0}69.90\textsubscript{7.00\phantom{0}} & \underline{\phantom{0}69.00\textsubscript{7.40\phantom{0}}} & \phantom{0}71.60\textsubscript{8.20\phantom{0}} & \phantom{0}77.50\textsubscript{8.70\phantom{0}} & \phantom{0}78.30\textsubscript{10.90} & \phantom{0}76.00\textsubscript{8.40\phantom{0}} & \phantom{0}69.60\textsubscript{7.30\phantom{0}} & \phantom{0}71.60\textsubscript{8.00\phantom{0}} & \phantom{0}71.40\textsubscript{8.80\phantom{0}} & \textbf{\phantom{0}12.10\textsubscript{1.80\phantom{0}}} \\
\bottomrule
\end{tabular}}
\caption{Comparisons of NEAURC of SGA and its contenders across 27 datasets for Sundial, where bold and underlined values denote the best and second-best results, respectively.}
\label{tab:all_comparisons_10}
\end{table*}

\begin{table*}[ht]
\centering
\footnotesize
\setlength{\tabcolsep}{1pt}
\renewcommand{\arraystretch}{0.78}
\resizebox{\linewidth}{!}{
\begin{tabular}{lcccccccccc}
\toprule
Dataset & NC & Ppl & PE & Eig & Ecc & Deg & SD & SAR & SE & SGA \\
\midrule
Australian Electricity & 173.40\textsubscript{26.00} & \underline{152.40\textsubscript{43.60}} & 154.60\textsubscript{40.40} & 152.40\textsubscript{43.10} & 152.40\textsubscript{43.00} & 152.40\textsubscript{42.60} & 152.40\textsubscript{42.30} & 152.40\textsubscript{47.70} & 154.60\textsubscript{41.50} & \textbf{\phantom{0}26.60\textsubscript{25.30}} \\
CIF 2016 & 130.70\textsubscript{16.50} & 124.90\textsubscript{12.50} & 129.80\textsubscript{13.90} & 119.60\textsubscript{16.60} & \underline{103.90\textsubscript{18.80}} & 119.70\textsubscript{17.30} & 113.50\textsubscript{17.60} & 122.20\textsubscript{15.90} & 131.70\textsubscript{15.90} & \textbf{\phantom{0}69.10\textsubscript{9.80\phantom{0}}} \\
Car Parts & 185.30\textsubscript{9.40\phantom{0}} & 178.90\textsubscript{8.00\phantom{0}} & 174.40\textsubscript{7.70\phantom{0}} & 165.60\textsubscript{7.40\phantom{0}} & \underline{143.50\textsubscript{7.20\phantom{0}}} & 165.80\textsubscript{7.60\phantom{0}} & 170.30\textsubscript{6.90\phantom{0}} & 168.10\textsubscript{8.10\phantom{0}} & 175.10\textsubscript{7.80\phantom{0}} & \textbf{\phantom{0}61.00\textsubscript{2.10\phantom{0}}} \\
Covid Deaths & \underline{106.00\textsubscript{15.30}} & 106.80\textsubscript{15.00} & 117.80\textsubscript{15.60} & 120.80\textsubscript{16.70} & 119.00\textsubscript{15.50} & 120.90\textsubscript{16.70} & 117.20\textsubscript{16.40} & 121.60\textsubscript{17.00} & 119.00\textsubscript{16.80} & \textbf{\phantom{0}80.60\textsubscript{9.80\phantom{0}}} \\
\bottomrule
\end{tabular}}
\caption{Comparisons of NEAURC of SGA and its contenders across 27 datasets for Aurora, where bold and underlined values denote the best and second-best results, respectively.}
\label{tab:all_comparisons_11}
\end{table*}